\documentclass{bmcart}
\usepackage{amsthm,amsmath}
\RequirePackage{natbib}
\RequirePackage{hyperref}
\usepackage[utf8]{inputenc} 
\usepackage[fleqn]{amsmath}
\usepackage{amsthm}
\usepackage{amssymb}
\usepackage{graphicx}
\usepackage{ragged2e}
\usepackage{geometry}
\usepackage{natbib}
\usepackage{algorithm}
\usepackage{algpseudocode}
\usepackage{chngcntr}
\newcounter{mycount}
\counterwithin{algorithm}{mycount}
\refstepcounter{mycount}

\usepackage[english]{babel}
\newtheorem{theorem}{Theorem}
\usepackage{float}
\newtheorem{remark}{Remark}
\newtheorem{corollary}{Corollary}
\newtheorem{definition}{Definition}
\startlocaldefs
\endlocaldefs

\begin{document}

\begin{frontmatter}

\begin{fmbox}
\dochead{Research}


\title{Semisupervised regression in latent structure networks on unknown manifolds}


\author[
   addressref={aff1},
   corref={aff1},
   email={aachary6@jh.edu}   
]{\inits{AA}\fnm{Aranyak} \snm{Acharyya}}
\author[
   addressref={aff2},
   email={jagterberg@jhu.edu}
]{\inits{JA}\fnm{Joshua} \snm{Agterberg}}
\author[
   addressref={aff3},
   email={mtrosset@indiana.edu}
]{\inits{MWT}\fnm{Michael} \snm{W. Trosset}}
\author[
   addressref={aff4},
   email={youngser@jhu.edu}
]{\inits{YP}\fnm{Youngser} \snm{Park}}
\author[
   addressref={aff1},
   email={cep@jhu.edu}
]{\inits{CEP}\fnm{Carey} \snm{E. Priebe}}


\address[id=aff1]{
  \orgname{Department of Applied Mathematics and Statistics,
  Johns Hopkins University,} 
  \city{Baltimore},                              
  \cny{USA}                                    
}
\address[id=aff2]{%
  \orgname{Institute for Data Engineering and Science, Department of Electrical and Systems Engineering, and Department of Statistics and Data Science,
  University of Pennsylvania,}
  \city{Pennsylvania},
  \cny{USA}
}
\address[id=aff3]{%
  \orgname{Department of Statistics,
  Indiana University,}
  \city{Bloomington},
  \cny{USA}
}
\address[id=aff4]{%
  \orgname{Centre for Imaging Science,
  Johns Hopkins University,}
  \city{Baltimore},
  \cny{USA}
}


\begin{artnotes}
\end{artnotes}

\end{fmbox}


\begin{abstractbox}

\begin{abstract} 
Random graphs are increasingly becoming objects of interest for modeling networks in a wide range of applications. Latent position random graph models posit that each node is associated with a latent position vector, and that these vectors follow some geometric structure in the latent space. In this paper, we consider random dot product graphs, in which an edge is formed between two nodes with probability given by the inner product of their respective latent positions. 
We assume that the latent position vectors lie on an unknown one-dimensional curve and are coupled with a response covariate via a regression model. Using the geometry of the underlying latent position vectors, we propose a manifold learning and graph embedding technique to predict the response variable on out-of-sample nodes, 
and we establish convergence guarantees for these responses. Our theoretical results are supported by simulations and an application to Drosophila brain data.
\end{abstract}


\begin{keyword}
\kwd{network inference}
\kwd{vertex covariates}
\kwd{random dot product graph}
\kwd{manifold learning}
\kwd{regression}
\end{keyword}


\end{abstractbox}
%

\end{frontmatter}



\section{Introduction}
\label{Intro}
\begin{figure}[!ht]
    \centering
    \includegraphics{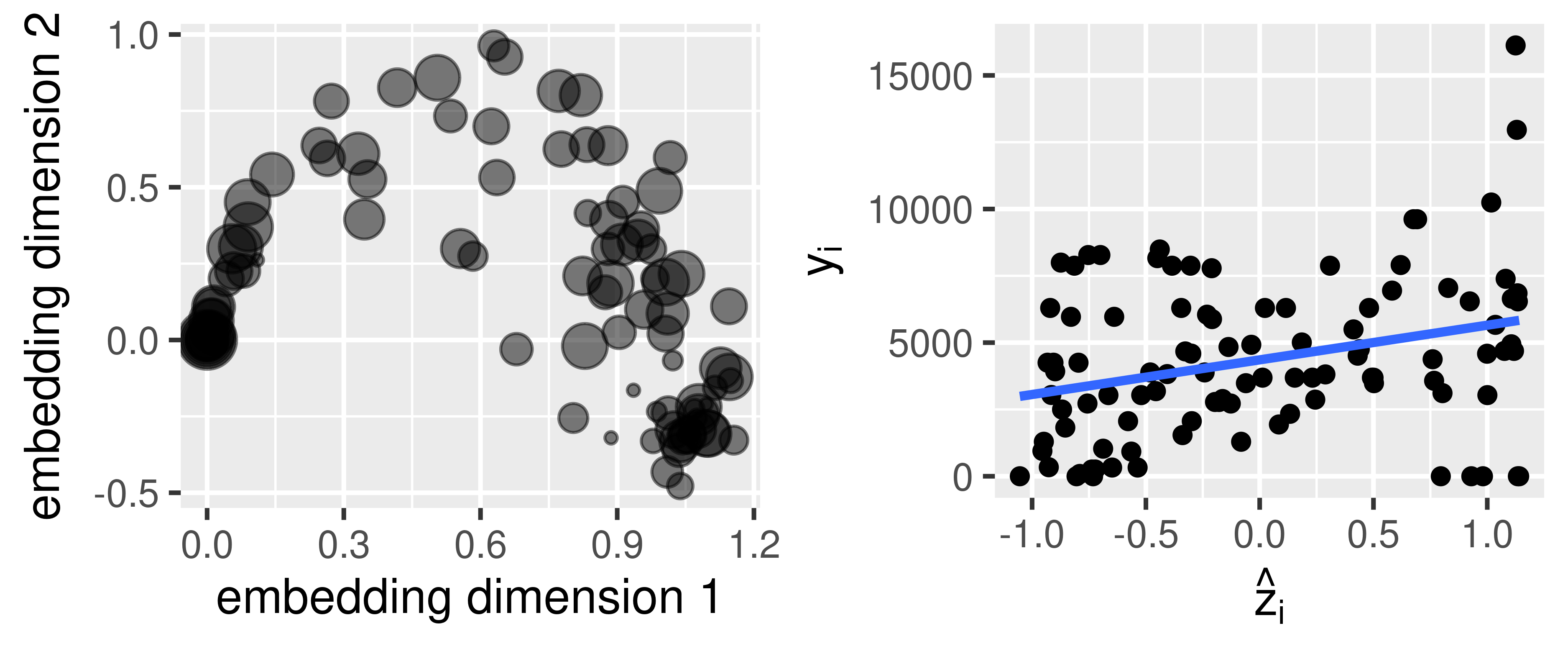}
    \caption{
Illustrative application of response prediction in latent structure networks on unknown manifolds.
Our methodology is
applied to the connectome of the right hemisphere of the Drosophila larval mushroom body. 
Left panel: scatter plot of two dimensions of the estimated latent positions for the 100 Kenyon cell neurons, obtained from spectral embedding of the network;
the dot size represents the response variable $y_i$ (the distance in microns between bundle entry point of neuron $i$ and the mushroom body neuropil).
Right panel: plot of responses $y_i$ against learnt
$1$-d embeddings $\hat{z}_i$ approximating
geodesic distances along this curve, for the 100 Kenyon cell neurons, together with the regression line.
In the left panel we observe that a one-dimensional curve captures nonlinear structure in the spectral embedding.
In the right panel we observe that response regressed against geodesic distance indicates a significant effect ($p < 0.01$ for $H_0: a=0$ in $y_i=a\hat{z}_i+b+\eta_i$).
}
\label{fig 1}
\end{figure}

Random graphs have long been an area of interest for scientists from different disciplines, primarily because of their applicability in modeling networks (\cite{Erdos1984OnTE},\cite{goldenberg2010survey}).
Latent position random graphs (\cite{Hoff2002LatentSA}) constitute a  category of random graphs where each node is associated with an unobserved vector, known as the latent position. One popular model, the
random dot product graph model (\cite{Young2007RandomDP}), comprise a subcategory of network models where the probability of edge formation between a pair of nodes is given by the inner product of their respective latent position vectors. 
This model was further generalized to the
generalized random dot product graph model (\cite{RubinDelanchy2022ASI})  which replaces the inner product with the indefinite inner product (see \cite{RubinDelanchy2022ASI}).
A survey of inference problems under the random dot product model can be found in
\cite{athreya2017statistical}. 
In \cite{rubin2020manifold}, it is shown that under certain regularity conditions, latent position random graphs can be equivalently thought of as generalized random dot product graphs whose nodes lie on a low dimensional manifold, which motivates the model we study in this work. Consider observing a random dot product graph whose latent positions lie on an unknown one-dimensional manifold in ambient space $\mathbb{R}^d$, and suppose
responses are recorded at some of these nodes. We choose to work in a semisupervised setting because in realistic scenarios, collecting observations is easier than obtaining labels corresponding to those observations. It is assumed that the responses are linked to the scalar pre-images of the corresponding latent positions via a regression model. In this \emph{semisupervised} setting, we aim to predict the responses at the out-of-sample nodes. 
\newline
\newline
The semisupervised learning framework in network analysis problems has been considered in a number of previous works.
In \cite{1326716}, a framework for regularization on graphs with labeled and unlabeled nodes was developed to predict the labels of unlabeled nodes. A dimensionality reduction technique was proposed from a graph-based algorithm developed to represent data on low dimensional manifold in high dimensional ambient space in \cite{6789755}. 
In the context of latent position networks with underlying 
low dimensional manifold structure, 
\cite{athreya2021estimation} discusses the problem of carrying out inference on the distribution of the latent positions of a random dot product graph, which are assumed to lie on a known low dimensional manifold in a  high dimensional ambient space. 
Moreover,
\cite{trosset2020learning} studies the problem of two-sample hypothesis testing for equality of means in a random dot product graph whose latent positions lie on a one-dimensional manifold in a high dimensional ambient space, where the manifold is unknown and hence must be estimated. To be more precise, 
\cite{trosset2020learning} proposes a methodology to learn the underlying manifold, and proves that the power of the statistical test based on the resulting embeddings can approach the power of the test based on the knowledge of the true manifold.
\newline
\newline
In our paper, we study the problem of predicting response covariate in a semisupervised setting, in a random dot product graph whose latent positions lie on an unknown one-dimensional manifold in  ambient space $\mathbb{R}^d$. Our main result establishes a convergence guarantee for the predicted responses when the manifold is learned using a particular manifold learning procedure (see \textit{Section \ref{Subsec:manifold_learning}}). As a corollary to our main result, we derive  a convergence guarantee for the power of the test for model validity based on the resulting embeddings. To help develop intuition, we first consider the problem of regression parameter estimation assuming the underlying manifold is known, and we show that a particular estimator is consistent in this setting.
\newline
\newline
We present an illustrative example of an application of our theoretical results. 
A connectome dataset consisting of a network of $100$ Kenyon cell neurons in larval \textit{Drosophila} (details in \cite{Eichler2017TheCC}) indicates the presence of an underlying low dimensional manifold structure. Each node
(that is, each Kenyon cell) is coupled with a response covariate, and the latent position of each node is estimated by a six-dimensional vector, using adjacency spectral embedding (see \textit{Section \ref{Subsec:RDPG}}). A scatterplot is obtained for each pair of dimensions of the estimated latent positions, and thus a $6 \times 6$ matrix of scatterplots is obtained (\textit{Figure \ref{fig 6}}). Each dimension is seen to be approximately related to another, and hence it is assumed that the latent positions lie on an one-dimensional manifold in six-dimensional ambient space. 
In order to capture the underlying structure, we construct a localization graph on the estimated latent positions and embed the dissimilarity matrix of shortest path distances into one-dimension 
(see \textit{Section \ref{Subsec:manifold_learning}}
for description of the method of embedding).
A scatterplot of the first two dimensions of the estimated latent positions is presented in \textit{Figure \ref{fig 1}}, where the size of the points varies as per the values of the associated response  covariate. 
A scatterplot of the responses $y_i$ against the one-dimensional  embeddings $\hat{z}_i$ is also presented along with fitted regression line indicating a significant effect. These results demonstrate that it will be reasonable to posit that the responses are 
linked to the embeddings via a simple linear regression model.
\newline
\newline
\newline
Our analysis raises the question of testing the validity of a simple linear regression model assumed to be linking the nodal responses to the scalar pre-images of the latent positions.  Our theory shows that the power of the test for validity of the model based on the raw-stress embeddings approximates the power of the test based on the true regressors.
\newline
\newline
In \textit{Section \ref{Background}} , we discuss key points about random dot product graphs and manifold learning.  \textit{Section \ref{Model_and_Methodology}} discusses the models and the methods to carry out subsequent inference on the corresponding regression models.
\textit{Section \ref{Theoretical_Results}} presents our theoretical results in both the settings of known and unknown manifolds.
\textit{Section \ref{Simu}} presents our findings from simulations.  \textit{Section \ref{Appli}} revisits our connectome application. \textit{Section 
\ref{Conc}} discusses the results and poses some questions that require further investigation. The proofs of our theoretical results are given in \textit{Section \ref{Appendix}}.
\section{Background notations, definitions and results}
\label{Background}
In this section, we introduce and explain the notations used throughout the paper. We also state relevant definitions and results pertaining to random dot product graphs (in \textit{Section \ref{Subsec:RDPG}}) and manifold learning (in \textit{Section \ref{Subsec:manifold_learning}}). 
\subsection{Notations}
We shall denote a vector by a bold lower-case letter, $\mathbf{x}$ for instance. Bold upper-case letters such as $\mathbf{P}$ will be used to represent matrices. The Frobenius norm and the maximum row-norm of a matrix $\mathbf{B}$ will be denoted respectively by $\left\lVert \mathbf{B} \right\rVert_F$ and
$\left\lVert \mathbf{B} \right\rVert_{2,\infty}$. 
The $i$-th row of a matrix $\mathbf{B}$ will be denoted by $\mathbf{B}_{i*}$, and the $j$-th column of $\mathbf{B}$ will be denoted by $\mathbf{B}_{*j}$. We will denote the $n \times n$ identity matrix by
$\mathbf{I}_n$, and $\mathbf{J}_n$ will denote the $n \times n$ matrix whose each entry equals one. Also, $\mathbf{H}_n=\mathbf{I}_n-\frac{1}{n} \mathbf{J}_n$ will denote the $n \times n$ centering matrix.
Unless otherwise mentioned, $\left\lVert \mathbf{x} \right\rVert$ will represent the Euclidean norm of a vector $\mathbf{x}$. The set of all orthogonal $d \times d$ matrices will be denoted by $\mathcal{O}(d)$. The set of all positive integers will be denoted by $\mathbb{N}$ and for any $n \in \mathbb{N}$, $[n]$ will denote the set $\left\lbrace 1,2,3...,n \right\rbrace$.
\subsection{Preliminaries on Random Dot Product Graphs}
\label{Subsec:RDPG}
A graph is an ordered pair $(V,E)$ where $V$ is the set of vertices (or nodes) and $E \subset V \times V$ is the set of edges connecting vertices. An adjacency matrix $\mathbf{A}$ of a graph is defined as
$\mathbf{A}_{ij}=1$ if $(i,j) \in E$, and $\mathbf{A}_{ij}=0$ otherwise. Here, we deal with hollow and undirected graphs; hence $\mathbf{A}$ is symmetric and
$\mathbf{A}_{ii}=0$ for all $i$. Latent position random graphs are those for which each node is associated with a vector that is called its latent position, denoted by $\mathbf{x}_i$, and the probability of formation of an edge between the $i$-th and $j$-th nodes is given by $\kappa(\mathbf{x}_i,\mathbf{x}_j)$ where $\kappa$ is a suitable kernel. 
\newline
\newline
Random vectors drawn from any arbitrary probability distribution cannot be latent positions of a random dot product graph, as their magnitudes can be unbounded whereas probabilities must lie in the interval $[0,1]$. The following definition allows us to work with a restricted class of distributions more amenable to random dot product graphs.
\begin{definition}
\textbf{(Inner product distribution)}:
\label{Def1:Inprod}
If  $F$ is a probability distribution function on $\mathbb{R}^d$ such that for any $\mathbf{x}, \mathbf{y} \in \mathrm{supp}(F)$, $\mathbf{x}^T \mathbf{y} \in [0,1]$, then $F$ is called an inner product distribution on $\mathbb{R}^d$.
\end{definition}
Next, we define the random dot product graphs, the basis of the models considered in this paper.
\begin{definition}
    (\textbf{Random Dot Product Graph}):
\label{Def2:RDPG}
Suppose $G$ is a hollow, undirected random graph with latent positions
$\mathbf{x}_1, \dots \mathbf{x}_n \in \mathbb{R}^d$. Let $\mathbf{X}=[\mathbf{x}_1| \dots |\mathbf{x}_n]^T$ be its latent position matrix and $\mathbf{A}$ be its adjacency matrix.
The graph $G$ is called random dot product graph if for all $i<j$, $\mathbf{A}_{ij} \sim \mathrm{Bernoulli}(\mathbf{x}_i^T \mathbf{x}_j)$ independently. The probability distribution of $\mathbf{A}$ is given by
\begin{displaymath}
P[\mathbf{A}]=\Pi_{i<j} (\mathbf{x}_i^T \mathbf{x}_j)^{\mathbf{A}_{ij}}
(1-\mathbf{x}_i^T \mathbf{x}_j)^{1-\mathbf{A}_{ij}}.
\end{displaymath}
If $\mathbf{x}_1,...\mathbf{x}_n \sim^{iid} F$ are the latent positions   where $F$ is an inner product distribution, then we write $(\mathbf{A},\mathbf{X}) \sim RDPG(F)$.
The distribution of the adjacency matrix conditional upon $\mathbf{x}_1,...\mathbf{x}_n$, is
\begin{displaymath}
    P[\mathbf{A}|\mathbf{X}]= \Pi_{i<j} (\mathbf{x}_i^T \mathbf{x}_j)^{\mathbf{A}_{ij}}(1-\mathbf{x}_i^T \mathbf{x}_j)^{1-\mathbf{A}_{ij}}.
\end{displaymath}
\end{definition}
The latent positions of a random dot product graph are typically unknown and need to be estimated in practice. The following definition puts forth an estimate of these latent positions via adjacency spectral embedding. 
\begin{definition}
(\textbf{Adjacency spectral embedding}):
\label{Def4: ASE}
 Suppose $\lambda_i$ is the $i$-th largest (in magnitude) eigenvalue of $\mathbf{A}$, and let $\mathbf{v}_i$ be the corresponding orthonormal eigenvector. Define 
 $\mathbf{S}=\mathrm{diag}(\lambda_1, \dots \lambda_d)$ and
 $\mathbf{V}=[\mathbf{v}_1| \dots | \mathbf{v}_d]$.
We define $\hat{\mathbf{X}}$, the adjacency spectral embedding of $\mathbf{A}$ into $\mathbb{R}^d$, via
$\hat{\mathbf{X}}=\mathbf{V}|\mathbf{S}|^{\frac{1}{2}}$.
\end{definition}
Now, we present two results from the literature which give us the consistency and asymptotic normality of suitably rotated adjacency spectral estimates of the latent positions of a random dot product graph.
\newline
\textit{
\begin{theorem}
\label{Th_ASE_const}
(\textit{Theorem $8$} from \cite{athreya2017statistical}): Suppose $\mathbf{x}_1,\mathbf{x}_2,...\mathbf{x}_n \in \mathbb{R}^d$ denote the latent positions of a random dot product graph with $n$ nodes, and let $\mathbf{X}^{(n)}=[\mathbf{x}_1|\mathbf{x}_2|...|\mathbf{x}_n]^T$ be the latent position matrix. Assume that for all $n$ sufficiently large, $rank(\mathbf{X}^{(n)})=d$, and there exists $a>0$ such that 
$b_n=\max_{i \in [n]} \sum_{j=1}^d \mathbf{x}_i^T \mathbf{x}_j > (\log(n))^{4+a}$. Suppose $\mathbf{A}^{(n)}$ denotes the corresponding adjacency matrix and let $\hat{\mathbf{X}}^{(n)}$ be the adjacency spectral embedding of $\mathbf{A}^{(n)}$ in $\mathbb{R}^d$ . There exists
a constant $C>0$ and
a sequence $\mathbf{W}^{(n)} \in \mathcal{O}(d)$  such that 
\begin{equation}
\lim_{n \to \infty} \mathbb{P}
\left[ 
\max_{i \in [n]}  
\left\lVert
(\hat{\mathbf{X}}^{(n)} \mathbf{W}^{(n)}
- \mathbf{X}^{(n)})_{i*} 
\right\rVert
\leq 
\frac
{C (\log(n))^2}
{\sqrt{b_n}}
\right]=1
\label{eq:Th1st1}
\end{equation}
Observe that under the assumption $b_n>(\log(n))^{4+a}$
, Eq. (\ref{eq:Th1st1}) implies
\newline
$\left\lVert \hat{\mathbf{X}}^{(n)} \mathbf{W}^{(n)}
- \mathbf{X}^{(n)} \right\rVert_{2,\infty} \to^P 0$ as $n \to \infty$.
\end{theorem}
}
Henceforth, we will denote this optimally rotated adjacency spectral embedding by $\Tilde{\mathbf{X}}^{(n)}$, that is, 
$\Tilde{\mathbf{X}}^{(n)}=\hat{\mathbf{X}}^{(n)} \mathbf{W}^{(n)}$. To simplify notations we will often omit the superscript $n$, and we will use 
$\Tilde{\mathbf{x}}_i$ to denote the $i$-th row of $\Tilde{\mathbf{X}}$.
Observe that in the attempt to estimate the true latent positions from the adjacency matrix, we encounter an inherent non-identifiability issue: for any  $\mathbf{W} \in \mathcal{O}(d)$, 
$\mathbb{E}(\mathbf{A}|\mathbf{X})=\mathbf{X} \mathbf{X}^T=(\mathbf{X}\mathbf{W}) 
(\mathbf{X}\mathbf{W})^T
$. This is the reason why the adjacency spectral embedding  needs to be rotated suitably so that it can approximate the true latent position matrix.
\textit{
\begin{theorem} 
\label{Th_ASE_asy}
(\textit{Theorem $9$} from \cite{athreya2017statistical}): Suppose $(\mathbf{A}^{(n)}, \mathbf{X}^{(n)}) \sim RDPG(F)$ be  adjacency matrices and latent position matrices  for a sequence of random dot product graphs, for which the latent positions are generated from an inner product distribution $F$ on $\mathbb{R}^d$.
Let
\begin{equation}
    \mathbf{\Sigma}(\mathbf{x_0})=
    \Delta^{-1} 
    \mathbb{E}_{\mathbf{x} \sim F} 
    \left[
    \mathbf{x_0}^T \mathbf{x}(1-\mathbf{x_0}^T \mathbf{x})
    \mathbf{x} \mathbf{x}^T
    \right] \Delta^{-1}
\label{eq:Th2st1}
\end{equation}
where $\Delta=\mathbb{E}_{\mathbf{x} \sim F}[\mathbf{x} \mathbf{x}^T]$. 
Then, there exists a sequence $\mathbf{W}^{(n)} \in \mathcal{O}(d)$  such that
for all $\mathbf{u} \in \mathbb{R}^d$,
\begin{equation}
\lim_{n \to \infty}
\mathbb{P}
\left[
\sqrt{n} (\hat{\mathbf{X}}^{(n)} \mathbf{W}^{(n)} -
\mathbf{X}^{(n)})_{i*} \leq \mathbf{u}
\right] = \int_{\mathbf{x} \in supp(F)}
\Phi_{\mathbf{0},\mathbf{\Sigma}(\mathbf{x})} (\mathbf{u}) dF(\mathbf{x})
\label{eq:Th2st2}
\end{equation}
where $\Phi_{\mathbf{0},\mathbf{\Sigma}(\mathbf{x})}(.)$ denotes the distribution function of multivariate normal $N(\mathbf{0},\mathbf{\Sigma}(\mathbf{x}))$ distribution.
\end{theorem}
}
Under suitable regularity conditions, a combined use of \textit{Theorem \ref{Th_ASE_asy}} and Delta method gives us the asymptotic distribution of $\sqrt{n}(\gamma(\Tilde{\mathbf{x}}_i)-\gamma(\mathbf{x}_i))$ for a function $\gamma:\mathbb{R}^d \to \mathbb{R}$, which depends on the true distribution $F$ of the latent positions. Therefore,  $\mathrm{var}(\gamma(\Tilde{\mathbf{x}}_i)-\gamma(\mathbf{x}_i))$ can be approximated from the optimally rotated adjacency spectral estimates $\Tilde{\mathbf{x}}_i$ and their empirical distribution function. 
In a random dot product graph for which the latent positions lie on a known one-dimensional manifold in ambient space $\mathbb{R}^d$, and the nodal responses are linked to the scalar pre-images  of the latent positions via a simple linear regression model, 
we can use the approximated variance of $(\gamma(\Tilde{\mathbf{x}}_i)-\gamma(\mathbf{x}_i))$ (motivated by works in \textit{Chapters $2$, $3$} of
\cite{fuller1987measurement}) to improve the performance (in terms of mean squared errors) of the naive estimators of the regression parameters, which are obtained by replacing $\mathbf{x}_i$ with $\Tilde{\mathbf{x}}_i$ in the least square estimates of the regression parameters. We demonstrate this method in detail in \textit{Section \ref{Conc}} (see \textit{Figure \ref{fig 8}}).
\newline
\newline
\begin{remark}
 Suppose
$F$ is a probability distribution satisfying $\mathrm{supp}(F)=K \subset \mathbb{R}^d$, and
$\mathbf{z}_1,....\mathbf{z}_n \sim^{iid} F$  are latent positions of a hollow symmetric latent position random graph with associated kernel $\kappa$. Extensive works presented in \cite{rubin2020manifold} show that if $\kappa \in L^2(\mathbb{R}^d \times \mathbb{R}^d)$, then there exists a mapping $q: \mathbb{R}^d \to L^2(\mathbb{R}^d)$ such that the graph can be equivalently represented as a generalized random dot product graph with latent positions $\mathbf{x}_i=q(\mathbf{z}_i) \in L(\mathbb{R}^d)$. If $\kappa$ is assumed to be H\"older continuous with exponent $c$, then
the Hausdorff dimension of $q(K)$ can be bounded by $\frac{d}{c}$, as shown in \cite{rubin2020manifold}. In \cite{whiteley2022discovering},
it has been shown that if $K$ is a Riemannian manifold, then stronger assumptions lead us to the conclusion that $q(K) \subset L^2(\mathbb{R}^d)$ is also Riemannian manifold diffeomorphic to $K$. Thus, under suitable regularity assumptions, any latent position graph can be treated as a generalized random dot product graph with latent positions on a low dimensional manifold. 
\end{remark}
After stating the relevant definitions and results pertinent to random dot product graphs, in the following section
we introduce the manifold learning technique we will use in this paper. Just for the sake of clarity, the topic of manifold learning in general has nothing to do with random dot product graph model; hence \textit{Section \ref{Subsec:RDPG}}
and \textit{Section \ref{Subsec:manifold_learning}} can be read independently.
\subsection{Manifold learning by raw-stress minimization}
\label{Subsec:manifold_learning}
Our main model is based on a random dot product graph whose latent positions lie on a one-dimensional Riemannian manifold.
Since one-dimensional Riemannian manifolds are
isometric to one-dimensional Euclidean space, we wish to represent the latent positions as points on the real line. This is the motivation behind the use of the following manifold learning technique, which relies upon approximation of geodesics by shortest path distances on localization graphs (\cite{tenenbaum2000global}, 
\cite{bernstein2000graph}, \cite{trosset2021rehabilitating}).
Given points $\mathbf{x}_1, \dots \mathbf{x}_n \in \mathcal{M}$ where $\mathcal{M}$ is an unknown one-dimensional manifold in ambient space $\mathbb{R}^d$, the goal is to find
$\hat{z}_1, \dots \hat{z}_n \in \mathbb{R}$, such that the interpoint Euclidean distances between $\hat{z}_i$ approximately equal the interpoint geodesic distances between $\mathbf{x}_i$. 
However, the interpoint geodesic distances between $\mathbf{x}_i$ are unknown. The following result shows how to estimate these unknown geodesic distances under suitable regularity assumptions.
\begin{theorem}
(\textit{Theorem $3$} from 
\cite{trosset2021rehabilitating})
\label{Th_man:geodesic_approx}
Suppose $\mathcal{M}$ is a one-dimensional compact Riemannian manifold in ambient space $\mathbb{R}^d$. Let $r_0$ and $s_0$ be the minimum radius of curvature and the minimum branch separation of $\mathcal{M}$.
Suppose $\nu$ is given and suppose $\lambda>0$ is chosen such that $\lambda<s_0$ and $\lambda<\frac{2}{\pi}r_0 \sqrt{24 \nu}$. Additionally, assume $\mathbf{x}_1, \dots \mathbf{x}_n \in \mathcal{M}$ are such that for every $\mathbf{u} \in \mathcal{M}$,
$d_M(\mathbf{u},\mathbf{x}_i)<\delta$.
 A localization graph is constructed on $\mathbf{x}_i$ as nodes under the following rule: two nodes $\mathbf{x}_i$ and $\mathbf{x}_j$ are joined by an edge if $\left\lVert \mathbf{x}_i-\mathbf{x}_j \right\rVert<\lambda$. When
 $\delta<\frac{\nu \lambda}{4}$,
the following condition holds for all $i,j \in [n]$,
\begin{equation*}
    (1-\nu)d_M(\mathbf{x}_i,\mathbf{x}_j) \leq d_{n,\lambda}(\mathbf{x}_i,\mathbf{x}_j) \leq
    (1+\nu) d_M(\mathbf{x}_i,\mathbf{x}_j),
\end{equation*}
where $d_{n,\lambda}(\mathbf{x}_i,\mathbf{x}_j)$ denotes the shortest path distance between $\mathbf{x}_i$ and $\mathbf{x}_j$. 
\end{theorem}
Given the dissimilarity matrix $\mathbf{D}=\left(
d_{n,\lambda}(\mathbf{x}_i,\mathbf{x}_j)
\right)_{i,j=1}^n$, the raw-stress function at $(z_1,\dots z_n)$ is defined as 
\begin{equation*}
   \sigma(z_1,\dots z_n)=
   \sum_{i<j} w_{ij}
   (|z_i-z_j|-d_{n,\lambda}(\mathbf{x}_i,\mathbf{x}_j))^2
\end{equation*}
where $w_{ij} \geq 0$ are weights. 
For the purpose of learning the manifold $\mathcal{M}$, we set 
$w_{ij}=1$ for all $i,j$, and compute
\begin{equation*}
    (\hat{z}_1, \dots \hat{z}_n)
    =\arg \min \sigma(z_1, \dots z_n)
    =\arg \min \sum_{i<j} 
   (|z_i-z_j|-d_{n,\lambda}(\mathbf{x}_i,\mathbf{x}_j))^2.
\end{equation*}
Since the scalars $\hat{z}_i$ are obtained by embedding $\mathbf{D}$ into one-dimension upon minimization of raw-stress, we shall henceforth refer to $\hat{z}_i$ as the one-dimensional raw-stress embeddings of $\mathbf{D}$.
\begin{remark}
In practice, raw-stress is minimized numerically by iterative majorization (\textit{Chapter $8$ of \cite{borg2005modern}}). Standard algorithms can sometimes be trapped in nonglobal minima. However, nearly optimal values can be obtained by repeated iterations of Guttman transformation (\textit{Chapter $8$ of
\cite{borg2005modern}}) when the configurations are intialized by classical multidimensional scaling.
In our paper, for theoretical results, we assume that the global minima is achieved.  
\end{remark}
\section{Model and Methodology}
\label{Model_and_Methodology}
Here we describe our models under both the assumptions of known and unknown manifold.
In each case,
we assume that we observe a random dot product graph for which the latent positions of the nodes lie on a one-dimensional manifold in $d$-dimensional ambient space.
Under the assumption that the underlying manifold is known, each node is coupled with a response linked to the scalar pre-image of the corresponding latent position via a regression model, and our goal is to estimate the regression parameters. When the underlying manifold is assumed to be unknown, our model involves a network with a small number of labeled nodes and a large number of unlabeled nodes, and our objective is to predict the response at a given unlabeled node assuming that the responses for the labeled nodes are linked to the scalar pre-images of the respective latent positions via a regression model. In the setting of unknown manifold, we can approximate the true regressors only up to scale and location transformations, and due to this non-identifiablity issue we carry out prediction of responses instead of estimation of regression parameters when the manifold is unknown.  
\begin{remark}
    We would like to remind the reader here that the setting of the known manifold is not realistic. We take the setting of the known manifold into account to help familiarize the reader with the best case scenario, for sake of comparison with the results obtained in the realistic setting of the unknown manifold.
\end{remark}
\subsection{Regression parameter estimation on known manifold}
\label{Model_Kn_man}
Suppose $\psi:\mathbb{R} \to \mathbb{R}^d$ is a known bijective function and let $\mathcal{M}=\psi([0,L])$ be a one-dimensional compact Reimannian manifold.
Consider a random dot product graph for which the nodes (with latent 
positions $\mathbf{x}_i$) lie on the known one-dimensional manifold $\mathcal{M}$ in $d$-dimensional ambient space. Let $t_1, \dots t_n$ be the scalar pre-images of the latent positions such that $\mathbf{x}_i=\psi(t_i)$ for all $i \in [n]$, where $n$ is the number of nodes of the graph. Suppose, for each $i$, the $i$-th node is coupled with a response $y_i$ which is linked to the latent position via the following regression model
\begin{equation}
\begin{aligned}
    y_i=\alpha+\beta t_i +\epsilon_i, i \in [n]
\end{aligned}
\end{equation}
where $\epsilon_i \sim^{iid} N(0,\sigma^2_{\epsilon})$ for all $i \in [n]$. Our goal is to estimate $\alpha$ and $\beta$.
\newline
If the true regressors $t_i$ were known, we could estimate $\alpha$ and $\beta$ by their ordinary least square estimates given by
\begin{equation}
\begin{aligned}
\hat{\beta}_{true}
=\frac
{
\sum_{i=1}^n (y_i-\Bar{y})(t_i-\Bar{t})
}
{
\sum_{i=1}^n (t_i-\Bar{t})^2
}, \hspace{0.5cm}
\hat{\alpha}_{true}=\Bar{y}-\hat{\beta}_{true} \Bar{t}.
\end{aligned}
\end{equation}
Since the true latent positions $\mathbf{x}_i$ are unknown, we estimate the true regressors $t_i$ by
\newline
 $\hat{t}_i=\arg \min_t \left\lVert \Tilde{\mathbf{x}}_i-\psi(t) \right\rVert$ where $\Tilde{\mathbf{x}}_i$ is the optimally rotated adjacency spectral estimate for the $i$-th latent position $\mathbf{x}_i$.
 The existence of $\hat{t}_i$ is guaranteed by the compactness of the manifold $\mathcal{M}=\psi([0,L])$.
 We then substitute $t_i$ by $\hat{t}_i$ in 
 $\hat{\alpha}_{true}$ and $\hat{\beta}_{true}$ to obtain the substitute (or the plug-in estimators) given by
 \begin{equation}
\begin{aligned}
\hat{\beta}_{sub}
=\frac
{
\sum_{i=1}^n (y_i-\Bar{y})(\hat{t}_i-\Bar{\hat{t}})
}
{
\sum_{i=1}^n (\hat{t}_i-\Bar{\hat{t}})^2
}, \hspace{0.5cm}
\hat{\alpha}_{sub}=\Bar{y}-\hat{\beta}_{sub} \Bar{\hat{t}}.
\end{aligned}
 \end{equation}
 The steps to compute $\hat{\alpha}_{sub}$ and $\hat{\beta}_{sub}$ are formally stated in \textit{Algorithm \ref{Algo1Known}}.
\newline
\begin{algorithm}[H]
\caption{EST($ 
\mathbf{A} \in \mathbb{R}^{n \times n}, \mathbf{W} \in \mathcal{O}(d),
d,\psi:[0,L] \to \mathbb{R}^d,
\left\lbrace y_i \right\rbrace_{i=1}^n
$)}
\label{Algo1Known}
\begin{algorithmic}[1]
\State Compute adjacency spectral embedding $\hat{\mathbf{X}}$ of the  adjacency matrix $\mathbf{A}$ into 
$\mathbb{R}^d$.
\State Use the given rotation matrix $\mathbf{W}$ to get the optimally rotated adjacency spectral embedding
$\Tilde{\mathbf{X}}=
\hat{\mathbf{X}} \mathbf{W}$.
\State Obtain the pre-images of the projections of the estimated latent positions on the manifold by
\newline
$\hat{t}_i=\arg \min_t \left\lVert \Tilde{\mathbf{x}}_i-\psi(t) \right\rVert$.
\State Compute the substitute estimators given by
$
\hat{\beta}_{sub}=\frac
    {
    \sum_{i=1}^n (y_i-\Bar{y})(\hat{t}_i-\Bar{\hat{t}})
    }
    {
    \sum_{i=1}^n (\hat{t}_i-\Bar{\hat{t}})^2
    }, \hspace{0.2cm}
    \hat{\alpha}_{sub}=\Bar{y}-\hat{\beta}_{sub} \Bar{\hat{t}}.
$
\State \Return $(\hat{\alpha}_{sub},\hat{\beta}_{sub})$.
\end{algorithmic}
\end{algorithm}
\subsection{Prediction of responses on unknown manifold}
Here, we assume that $\psi:[0,L] \to \mathbb{R}^d$ is unknown and arclength parameterized, that is, $\left\lVert \Dot{\psi}(t) \right\rVert=1$ for all $t$. 
Additionally, assume that $\mathcal{M}=\psi([0,L])$ is a compact Reimannian manifold.
Consider a $n$-node random dot product graph whose nodes (with latent positions $\mathbf{x}_i$) lie on the unknown manifold $\mathcal{M}=\psi([0,L])$ in ambient space $\mathbb{R}^d$. Assume that the first $s$ nodes of the graph are coupled with a response covariate and the response $y_i$ at the $i$-th node is linked to the latent position via a linear regression model
\begin{equation}
\begin{aligned}
y_i=\alpha+\beta t_i +\epsilon_i, i \in [s]
\end{aligned}
\end{equation}
where $\epsilon_i \sim^{iid} N(0,\sigma^2_{\epsilon})$ for all $i \in [s]$. 
Our goal is to predict the response for the $r$-th node, where $r>s$. 
First, we compute the adjacency spectral estimates $\hat{\mathbf{x}}_i$ of the latent positions of all $n$ nodes. We then construct a localization graph on the adjacency spectral estimates 
$\hat{\mathbf{x}}_i$ under the following rule: join two nodes $\hat{\mathbf{x}}_i$ and $\hat{\mathbf{x}}_j$ if and only if 
$\left\lVert \hat{\mathbf{x}}_i-\hat{\mathbf{x}}_j \right\rVert< \lambda$, for some pre-determined $\lambda>0$ known as the neighbourhood parameter. Denoting the shortest path distance between $\hat{\mathbf{x}}_i$ and $\hat{\mathbf{x}}_j$ by $d_{n,\lambda}(\hat{\mathbf{x}}_i,\hat{\mathbf{x}}_j)$, we embed the dissimilarity matrix $\mathbf{D}=\left(d_{n,\lambda}(\hat{\mathbf{x}}_i,\hat{\mathbf{x}}_j)  \right)_{i,j=1}^l$ into one-dimension by minimizing the raw-stress criterion, thus obtaining 
\begin{equation}
\begin{aligned}
(\hat{z}_1,....\hat{z}_l)
= \text{arg min} \sum_{i=1}^l \sum_{j=1}^l (|z_i-z_j|-d_{n,\lambda}(\hat{\mathbf{x}}_i,\hat{\mathbf{x}}_j))^2
\end{aligned}
\end{equation}
where $l$ is such that $s<r<l \leq n$.
We then use a simple linear regression model on the bivariate data $(y_i,\hat{z}_i)_{i=1}^s$ to predict the response for $\hat{z}_r$ corresponding to the $r$-th node.
The abovementioned procedure to predict the response at the $r$-th node from given observations is formally described in 
\textit{Algorithm \ref{Algo2unknown}}.
\begin{algorithm}[H]
\caption{PRED($
\mathbf{A} \in \mathbb{R}^{n \times n},
d,\lambda,l, \left\lbrace y_i \right\rbrace_{i=1}^s,r
$)}
\label{Algo2unknown}
\begin{algorithmic}[1]
\State Obtain the adjacency spectral estimates $\hat{\mathbf{x}}_1 \dots \hat{\mathbf{x}}_n \in \mathbb{R}^d$ of the latent positions from the adjacency matrix $\mathbf{A}$.
\State Construct a localization graph with $\hat{\mathbf{x}}_i$ as vertices by the following rule: join two vertices $\hat{\mathbf{x}}_i,\hat{\mathbf{x}}_j$ if
and only if $\left\lVert \hat{\mathbf{x}}_i-\hat{\mathbf{x}}_j \right\rVert<\lambda$. 
\State For every $i,j \in [n]$, get shortest path distance $d_{n,\lambda}(\hat{\mathbf{x}}_i,\hat{\mathbf{x}}_j)$. 
\State Obtain
$(\hat{z}_1,....\hat{z}_l)
= \text{arg min} \sum_{i=1}^l \sum_{j=1}^l (|z_i-z_j|-d_{n,\lambda}(\hat{\mathbf{x}}_i,\hat{\mathbf{x}}_j))^2$.
\State Compute 
$\hat{b}=\frac
{
\sum_{i=1}^s (y_i-\Bar{y})(\hat{z}_i-\Bar{\hat{z}})
}
{
\sum_{i=1}^s (\hat{z}_i-\Bar{\hat{z}})^2
}$ and
$\hat{a}=\Bar{y}-\hat{b} \Bar{\hat{z}}$.
\State For $r>s$, compute $\Tilde{y}_r=\hat{a}+\hat{b} \hat{z}_r$.
\State \Return $\Tilde{y}_r$. 
\end{algorithmic}
\end{algorithm}
\section{Main results}
\label{Theoretical_Results}
In this section we present our theoretical results showing consistency of the estimators of the regression parameters on a known manifold, and convergence guarantees for the predicted responses based on the raw-stress embeddings on an unknown manifold. In the setting of unknown manifold, as a corollary to consistency of the predicted responses, we also derive a convergence guarantee for a test for validity of a simple linear regression model based on an approximate $F$-statistic.
\subsection{The case of known manifold}
\label{Kn_man}
Recall that we observe a random dot product graph with $n$ nodes for which the latent positions $\mathbf{x}_i$ lie on a one-dimensional manifold.
 Our following result 
 shows that we can consistently estimate $(\alpha,\beta)$ by 
$(\hat{\alpha}_{sub},\hat{\beta}_{sub})$.
\textit{
\begin{theorem}
\label{Th_kn_const}
Suppose $\psi:[0,L] \to \mathbb{R}^d$ is bijective, and its inverse $\gamma$ satisfies $\left\lVert \nabla\gamma(\mathbf{w}) 
\right\rVert<K$ 
for all $\mathbf{w} \in \psi([0,L])$,
for some $K>0$. Let $\mathbf{x}_i=\psi(t_i)$ be the latent position of the $i$-th node of a random dot product graph with $n$ nodes, and assume $y_i=\alpha+\beta t_i +\epsilon_i$, $\epsilon_i \sim^{iid} N(0,\sigma^2_{\epsilon})$ for all $i \in [n]$. Assume $\mathbf{x}_i \sim^{iid} F$ for all $i$ where $F$ is an inner product distribution on $\mathbb{R}^d$.  Let $\mathbf{X}^{(n)}=[\mathbf{x}_1| \dots |\mathbf{x}_n]$ be the latent position matrix and suppose $\hat{\mathbf{X}}^{(n)}$ is the adjacency spectral embedding of the adjacency matrix $\mathbf{A}^{(n)}$ into $\mathbb{R}^d$. Assume $\hat{\mathbf{W}}^{(n)}
=
\arg \min_{\mathbf{W} \in \mathcal{O}(d)}
\left\lVert 
\hat{\mathbf{X}}^{(n)} \mathbf{W}
- \mathbf{X}^{(n)}
\right\rVert_F
$
is known.
Then, as $n \to \infty$, we have
$\hat{\alpha}_{sub} \to^P \alpha$ 
and
$\hat{\beta}_{sub} \to^P \beta$,
where
$(\hat{\alpha}_{sub},\hat{\beta}_{sub})=\mathrm{EST}(\mathbf{A}^{(n)},d,\hat{\mathbf{W}}^{(n)},\psi, \left\lbrace y_i \right\rbrace_{i=1}^n)$ (see \textit{Algorithm \ref{Algo1Known}}).
\end{theorem}
}
A rough sketch of proof for \textit{Theorem \ref{Th_kn_const}}
is as follows. Note that 
$\hat{t}_i=\arg \min_t
\left\lVert
\Tilde{\mathbf{x}}_i-\psi(t)
\right\rVert$ where $\Tilde{\mathbf{x}}_i$ is the optimally rotated adjacency spectral estimate of $\mathbf{x}_i$,
and
$
t_i=\arg \min_t
\left\lVert
\mathbf{x}_i-\psi(t)
\right\rVert
$.  Recall that \textit{Theorem \ref{Th_ASE_const}}
tells us that
$\Tilde{\mathbf{x}}_i$ is consistent for $\mathbf{x}_i$. This enables us to prove that $\hat{t}_i$ is consistent for $t_i$ which ultimately leads us to the consistency of $\hat{\alpha}_{sub}$ and $\hat{\beta}_{sub}$, because $\hat{\alpha}_{sub}$ and $\hat{\beta}_{sub}$ are computed by replacing the true regressors $t_i$
with $\hat{t}_i$ in the expressions of $\hat{\alpha}_{true}$ and $\hat{\beta}_{true}$ which are consistent for $\alpha$ and $\beta$ respectively.
\textit{Theorem \ref{Th_kn_const}} gives us the consistency of the substitute  estimators of the regression parameters under the assumption of boundedness of the gradient of the inverse function. Since continuously differentiable functions have bounded gradients on compact subsets of their domain, we can apply \textit{Theorem \ref{Th_kn_const}} whenever $\gamma=\psi^{-1}$ can be expressed as a restriction to a function with continuous partial derivatives.
As a direct consequence of \textit{Theorem \ref{Th_kn_const}}, our next result demonstrates that the substitute estimators  have optimal asymptotic variance amongst all linear unbiased estimators.
\textit{
\begin{corollary}
\label{Cor_kn_BLUE}
Conditioning upon the true regressors $t_i$ in the setting of \textit{Theorem \ref{Th_kn_const}},
the following two conditions hold
\newline
(A)
$
\mathbb{E}(\hat{\alpha}_{sub}) \to \alpha,
\hspace{0.5cm} 
\mathbb{E}(\hat{\beta}_{sub}) \to \beta 
\text{ as $n \to \infty$}.
$,
\newline
(B)
For any two linear unbiased estimators $\Tilde{\alpha}$ and $\Tilde{\beta}$ and an arbitrary $\delta>0$, 
\newline
$
\mathrm{var}(\hat{\alpha}_{sub}) \leq \mathrm{var}(\Tilde{\alpha})+\delta, \hspace{0.5cm}
\mathrm{var}(\hat{\beta}_{sub}) \leq \mathrm{var}(\Tilde{\beta})+\delta
$ for sufficiently large $n$.
\end{corollary}
}
\subsection{The case of unknown manifold}
\label{Unkn_man}
Recall that our goal is to predict responses in a semisupervised setting on a random dot product graph on an unknown one-dimensional manifold in ambient space $\mathbb{R}^d$.  We provide justification for the use of
\textit{Algorithm \ref{Algo2unknown}} for this purpose, by showing that the predicted response
$\Tilde{y}_r$ at the $r$-th node
based on the raw-stress embeddings
approaches the predicted response  based on the true regressors $t_i$ as $n \to \infty$.
\newline
\newline
Intuition suggests that in order to carry out inference on the regression model, we must learn the unknown manifold $\mathcal{M}$.
We exploit the availability of  large number of unlabeled nodes whose latent positions lie on the one-dimensional manifold, to  learn the manifold. 
Observe that since the underlying manifold $\psi([0,L])$
is arclength parameterized, the geodesic distance between any two points on it is the same as the interpoint distance between their corresponding pre-images. Results from the literature (\cite{bernstein2000graph}) show that if an appropriate localization graph is constructed on sufficiently large number of points on a manifold, then the shortest path distance between two points  approximates the geodesic distance between those two points. Therefore, on a localization graph of an appropriately chosen neighbourhood parameter $\lambda$, constructed on the adjacency spectral estimates $\hat{\mathbf{x}}_1,\hat{\mathbf{x}}_2,...,\hat{\mathbf{x}}_n$, the shortest path distance $d_{n,\lambda}(\hat{\mathbf{x}}_i,\hat{\mathbf{x}}_j)$
between $\hat{\mathbf{x}}_i$ and $\hat{\mathbf{x}}_j$ is expected to approximate the geodesic distance $d_M(\mathbf{x}_i,\mathbf{x}_j)=|t_i-t_j|$.
Note that here is no need to rotate the adjacency spectral estimates $\hat{\mathbf{x}}_i$, because the shortest path distance is sum of Euclidean distances and Euclidean distances are invariant to orthogonal transformations.
Thus, when the dissimilarity matrix $\mathbf{D}=\left( d_{n,\lambda}(\hat{\mathbf{x}}_i,\hat{\mathbf{x}}_j) \right)_{i,j=1}^l$ is embedded into one-dimension by minimization of raw-stress, we obtain embeddings $\hat{z}_1, \dots \hat{z}_l$
 for which interpoint distance $|\hat{z}_i-\hat{z}_j|$ approximates the interpoint distance $|t_i-t_j|$. In other words, the estimated distances from the raw-stress embeddings applied to the adjacency spectral estimates of the latent positions approximate the true geodesic distances, which is demonstrated by the following result. This argument is the basis for construction of a sequence of predicted responses based on the raw-stress embeddings, which approach the predicted responses based on the true regressors as the number of auxiliary nodes go to infinity. 
 \newline
\textit{
\begin{theorem}
\label{Th_emb_interpoint_const}
 (\textit{Theorem 4} from \cite{trosset2020learning}): Suppose the function $\psi:[0,L] \to \mathbb{R}^d$ is such that $\left\lVert \Dot{\psi}(t) \right\rVert=1$ for all $t \in [0,L]$.
 Let $\mathbf{x}_i \in \mathbb{R}^d$ be the latent position for the $i$-th node of the random dot product graph, and $t_i \in \mathbb{R}$ be such that $\mathbf{x}_i=\psi(t_i)$ for all $i$,
and assume $t_i \sim^{iid} g$ where $g$ is an absolutely continuous probability density function satisfying $supp(g)=[0,L]$ and $g_{min}>0$.
Let $\hat{\mathbf{x}}_i$ be the adjacency spectral estimate of the true latent position $\mathbf{x}_i$ for all $i$. There exist sequences 
$\left\lbrace n_K \right\rbrace_{K=1}^{\infty}$
of number  of nodes and 
$\left\lbrace \lambda_K \right\rbrace_{K=1}^{\infty}$ of neighbourhood parameters satisfying $n_K \to \infty$, $\lambda_K \to 0$ as $K \to \infty$, such that for any fixed integer $l$ satisfying 
$s<l<n_K$ for all $K$,
\begin{equation}
   (|\hat{z}_i-\hat{z}_j| - |t_i-t_j|) \to^P 0, \text{ for all } i,j \in [l]
\label{eq:Th4st1}
\end{equation}
holds, where $(\hat{z}_1,....\hat{z}_l)$ is minimizer of the raw stress criterion, that is
\begin{equation}
(\hat{z}_1,....\hat{z}_l)
= \text{arg min} \sum_{i=1}^l \sum_{j=1}^l (|z_i-z_j|-d_{n_K,\lambda_K}(\hat{\mathbf{x}}_i,\hat{\mathbf{x}}_j))^2.
\label{eq:Th4st2}
\end{equation}
\end{theorem}
}
 \textit{Theorem \ref{Th_emb_interpoint_const}} shows that 
the one-dimensional raw-stress embeddings
$\hat{z}_1,\dots \hat{z}_l$ satisfy  $(|\hat{z}_i-\hat{z}_j| - |t_i-t_j|) \to 0$ as $K \to \infty$,
for all $i,j \in [l]$.
This means that for every $i \in [l]$, $\hat{z}_i$ approximates $t_i$ up to scale and location transformations. Since in simple linear regression the accuracy of the predicted response is independent of scale and location transformations, we can expect the predicted response at a particular node based on $\hat{z}_i$ to approach the predicted response based on the true regressors $t_i$. The following theorem, our key result in this paper, demonstrates that this is in fact the case.
\textit{
\begin{theorem}
\label{Th_reg_pred_const}
 Consider a  random dot product graph for which each node lies on an arclength parameterized one-dimensional manifold $\psi([0,L])$ where $\psi$ is unknown.
Let $\mathbf{x}_i=\psi(t_i)$ be the latent position of the $i$-th node for all $i$. Assume  $y_i=\alpha+\beta t_i + \epsilon_i$, $\epsilon_i \sim^{iid} N(0,\sigma^2_{\epsilon})$ for $i \in [s]$,
where $s$ is a fixed integer.
The predicted response at the $r$-th node based on the true regressors is 
$\hat{y}_r=\hat{\alpha}_{true}+\hat{\beta}_{true} t_r$. 
There exist sequences $n_K \to \infty$ of number of nodes and $\lambda_K \to 0$ of neighbourhood parameters such that for every $r>s$, 
$
|\hat{y}_r-\Tilde{y}_r^{(K)}| \to^P 0
$
as $K \to \infty$,
where
$\Tilde{y}_r^{(K)}=\mathrm{PRED}(\mathbf{A}^{(K)},d,\lambda_K,l,
\left\lbrace y_i \right\rbrace_{i=1}^s, r
)$ (see \textit{Algorithm \ref{Algo2unknown}}), $\mathbf{A}^{(K)}$ being the adjacency matrix when the number of nodes is $n_K$ and $l$ being a fixed natural number that satisfies $l>r>s$.
\end{theorem}
}
Recall that the validity of a simple linear regression model can be tested by an $F$-test, whose test statistic is dependent on the predicted responses based on the true regressors. Since we have a way to approximate the predicted responses based on the true regressors by predicted responses based on the raw-stress embeddings, we can also devise a test whose power approximates the power of the $F$-test based on the true regressors, as shown by our following result.
\textit{
\begin{corollary}
\label{Cor_test_conv}
In the setting of \textit{Theorem \ref{Th_reg_pred_const}},
suppose $\left\lbrace (\Tilde{y}_1^{(K)},\Tilde{y}_2^{(K)},....
\Tilde{y}_s^{(K)})
\right\rbrace_{K=1}^{\infty}$ is the sequence of vector of predicted responses at the first $s$ nodes of the random dot product graph, based on the raw-stress embeddings $\hat{z}_1,...,\hat{z}_s$.
Define
\begin{equation}
F^*=
(s-2)
\frac
{
\sum_{i=1}^s (\hat{y}_i-\Bar{y})^2
}
{
\sum_{i=1}^s (y_i-\hat{y}_i)^2
},
\hspace{1cm}
\hat{F}^{(K)}=
(s-2)
\frac
{
\sum_{i=1}^s (\Tilde{y}^{(K)}_i-\Bar{y})^2
}
{
\sum_{i=1}^s (y_i-\Tilde{y}^{(K)}_i)^2
}.
\label{Cor2:stat}
\end{equation}
Consider testing the null hypothesis 
$H_0:\beta=0$ against $H_1:\beta \neq 0$ in the absence of the true regressors $t_i$, and the decision rule is: reject $H_0$ in favour of $H_1$ at level of significance $\Tilde{\alpha}$ if
$\hat{F}^{(K)}>c_{\Tilde{\alpha}}$,
where $c_{\Tilde{\alpha}}$ is the $(1-\Tilde{\alpha})$-th quantile of $F_{1,s-2}$ distribution.
If the power of this test is denoted by $\hat{\pi}^{(K)}$, then $\lim_{K \to \infty} \hat{\pi}^{(K)}=\pi^*$, where $\pi^*$ is the power of the test for which the decision rule is to reject $H_0$ in favour of $H_1$ at level of significance $\Tilde{\alpha}$ if $F^*>c_{\Tilde{\alpha}}$.
\end{corollary} 
}
Thus, if one wants to perform a test for model validity in the absence of the true regressors $t_i$, then a test of approximately equal power, based on the raw-stress embeddings $\hat{z}_i$ for a graph of sufficiently large number of auxiliary nodes, can be used instead. 
\newline
\section{Simulation}
\label{Simu}
In this section, we present our simulation results demonstrating support for our theorems. We conducted simulations on $100$  Monte Carlo samples of
graphs on known and unknown one-dimensional manifolds. 
\subsection{The case of known manifold}
\begin{figure}[!ht]
    \centering
    \includegraphics{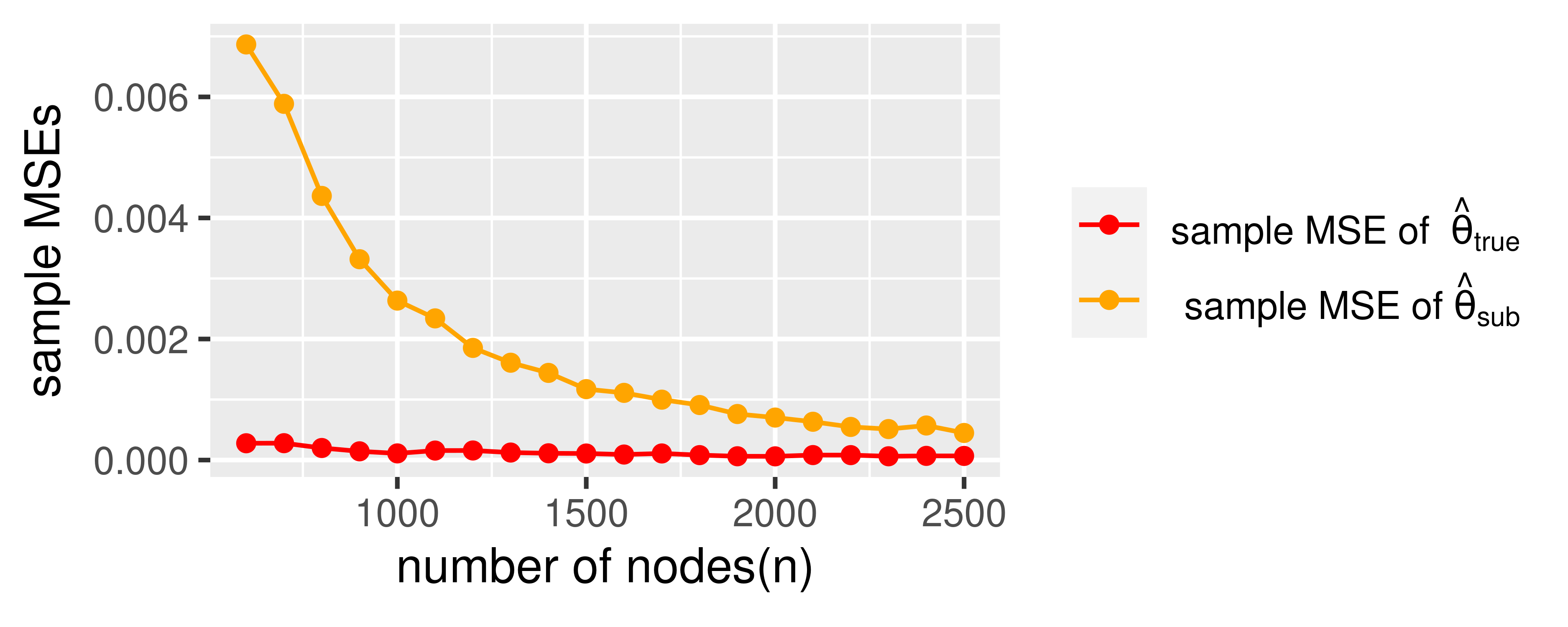}
    \caption{ Plot showing consistency of the substitute estimator of the regression parameter vector on known manifold. For $100$ Monte Carlo samples, substitute estimates are computed using the projections of the optimally rotated adjacency spectral estimates of the latent positions onto the manifold, and then the sample MSEs of the estimator based on the true regressors and the substitute estimator are computed.
    For graphs of moderate size ($n \leq 2000$), the substitute estimator performs significantly worse than the estimator based on the true regressors.
    However, as the number of nodes increases, the difference in performances of the estimators diminish and the mean squared errors of both the estimators approach zero.  
    }
    \label{fig 3}
\end{figure}

We take the manifold to be $\psi([0,1])$, where $\psi:[0,1] \to \mathbb{R}^3$ is the Hardy Weinberg curve, given by
$\psi(t)=(t^2,2t(1-t),(1-t)^2)$.
The number of nodes, $n$, varies from $600$ to $2500$ in steps of $100$. For each $n$, we repeat the following procedure over $100$ Monte Carlo samples.
 A sample $t_1,.....t_n \sim^{iid} U[0,1]$ is generated, and responses $y_i$ are sampled via the regression model $y_i=\alpha+\beta t_i+\epsilon_i$, $\epsilon_i \sim^{iid} N(0,\sigma^2_{\epsilon})$, $i \in [n]$, where $\alpha=2.0$, $\beta=5.0$, $\sigma_{\epsilon}=0.1$.
 An undirected hollow random dot product graph with latent positions $\mathbf{x}_i=\psi(t_i)$, $i \in [n]$ is generated. More specifically, the $(i,j)$-th element of the adjacency matrix $\mathbf{A}$ satisfies
$
\mathbf{A}_{ij} \sim \mathrm{Bernoulli}(\mathbf{x}_i^T \mathbf{x}_j)
$
for all $i<j$, and $\mathbf{A}_{ij}=\mathbf{A}_{ji}$ for all
$i,j \in [n]$, and $\mathbf{A}_{ii}=0$ for all $i \in [n]$.
We denote the true latent position matrix by $\mathbf{X}=[\mathbf{x}_1|\mathbf{x}_2|...|\mathbf{x}_n]^T$, and the adjacency spectral estimate of it by $\hat{\mathbf{X}}$. We compute
\begin{displaymath}
\hat{\mathbf{W}}=\arg \min_{\mathbf{W} \in \mathcal{O}(d)}
\left\lVert 
\mathbf{X}-\hat{\mathbf{X}} \mathbf{W}
\right\rVert_F
\end{displaymath}
and finally, set $\Tilde{\mathbf{X}}=\hat{\mathbf{X}} \hat{\mathbf{W}}$
to be the optimally rotated adjacency spectral estimate of the latent position matrix $\mathbf{X}$.
Then we obtain  $\hat{t}_i=\arg \min_t
\left\lVert
\Tilde{\mathbf{x}}_i-\psi(t)
\right\rVert
$
for $i \in [n]$,
and get 
$\hat{\alpha}_{sub}$ and $\hat{\beta}_{sub}$. Setting $\boldsymbol{\theta}=(\alpha,\beta)$, $\hat{\boldsymbol{\theta}}_{true}=(\hat{\alpha}_{true},\hat{\beta}_{true})$ and $\hat{\boldsymbol{\theta}}_{sub}=(\hat{\alpha}_{sub},\hat{\beta}_{sub})$, we compute the sample mean squared errors (MSE) of $\hat{\boldsymbol{\theta}}_{true}$ and
$\hat{\boldsymbol{\theta}}_{sub}$ over the $100$ Monte Carlo samples and plot them against $n$. The plot is given in \textit{Figure \ref{fig 3}}.
\newline
\newline
\begin{remark}
    The fact that the optimal rotation matrix $\hat{\mathbf{W}}$ needs to be computed from the true latent position matrix $\mathbf{X}$ is what makes inference on the regression model in the scenario of known manifold unrealistic, because $\mathbf{X}$ is typically unknown. 
\end{remark}
\subsection{The case of unknown manifold}
\begin{figure}[!ht]
    \centering
    \includegraphics{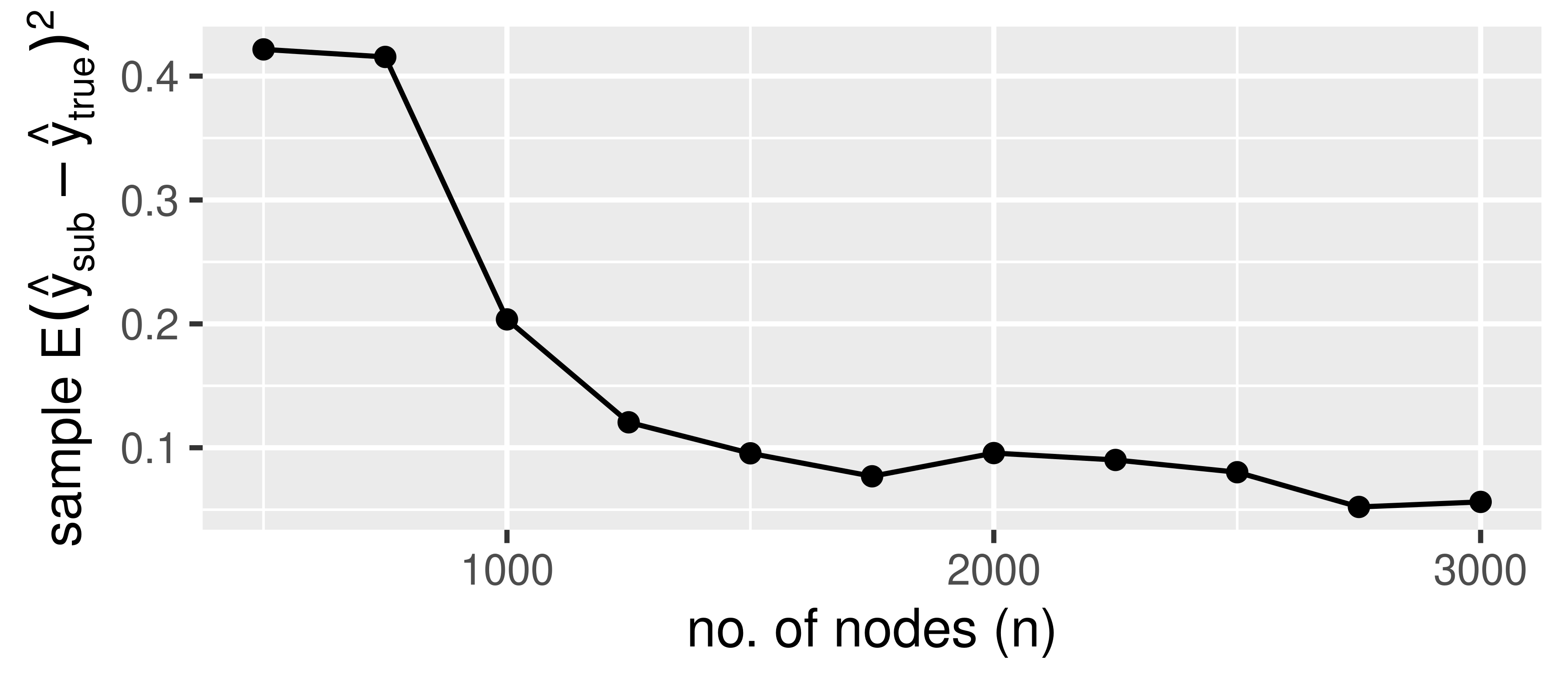}
    \caption{ Plot showing consistency of the predicted responses based on raw-stress embeddings on unknown manifold. The arclength parameterized manifold is taken to be $\psi([0,1])$ where $\psi(t)=(t/2,t/2,t/2,t/2)$. 
    The sample mean of the squared difference between the predicted response $\hat{y}_{sub}$ based on raw-stress embeddings and predicted response 
    $\hat{y}_{true}$
    based on the true regressors is plotted against the total number of nodes in the graph.
    The substitute predicted response $\hat{y}_{sub}$ performs significantly worse than $\hat{y}_{true}$ for moderate number of auxiliary nodes ($n \leq 1000$). 
     However, as the number of auxiliary nodes increases further,  the squared difference between $\hat{y}_{true}$ and $\hat{y}_{sub}$ goes to zero.  
    }
    \label{fig 4}
\end{figure}

\begin{figure}[!ht]
    \centering
\includegraphics{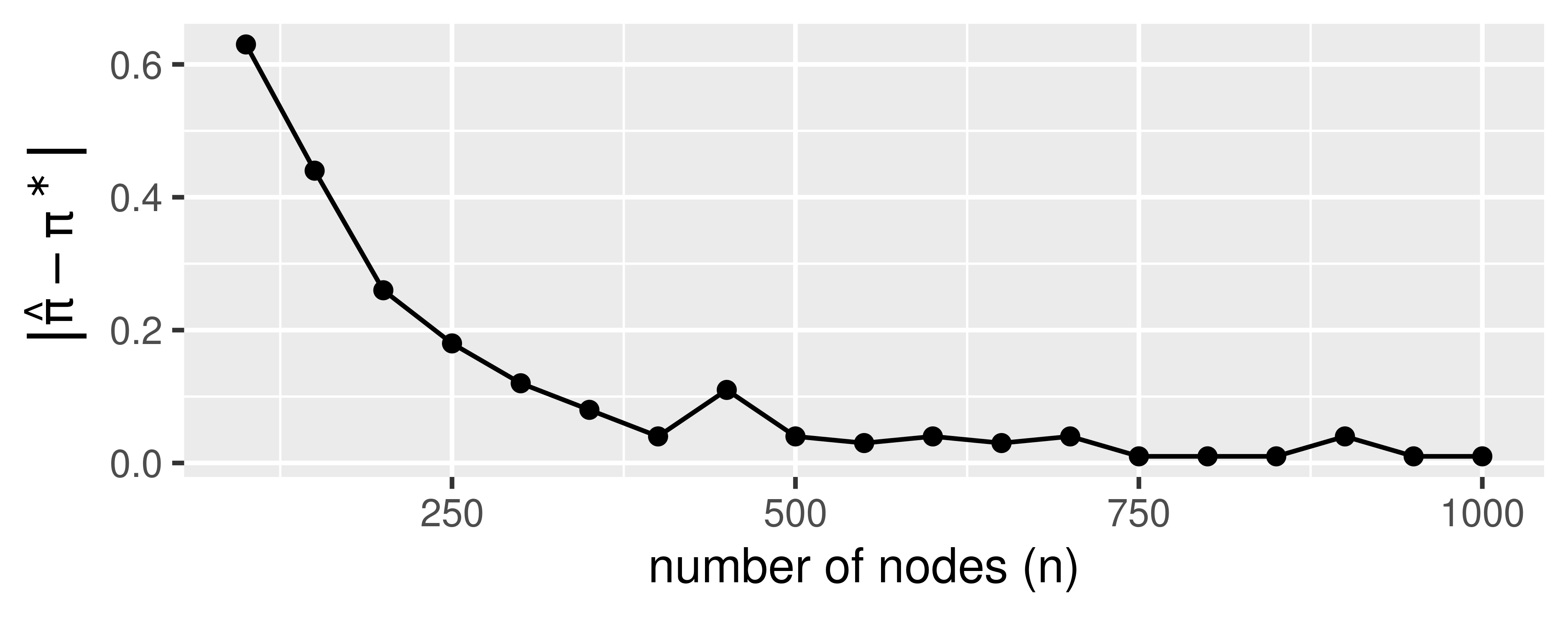}
    \caption{ Plot of the difference between the empirical powers of tests for model validity based on the $1$-dimensional raw-stress embeddings and the true regressors.
    The arclength parameterized manifold is taken to be $\psi([0,1])$ where $\psi(t)=(t/2,t/2,t/2,t/2)$. 
    For a small fixed number ($s=20$) of nodes, responses $y_i$ are generated from $y_i=\alpha+\beta t_i + \epsilon_i$,
    $\epsilon_i \sim^{iid} N(0,\sigma^2_{\epsilon})$.
    A large number $(n-s)$ of auxiliary nodes are generated on $\psi([0,1])$ and a localization graph is constructed on the adjacency spectral estimates corresponding to the first $n/2$ nodes. When $n$ is the $K$-th term of the vector $(100,150,200,\dots 1000)$, the neighbourhood parameter is taken to be $\lambda=0.9 \times 0.99^{K-1}$.
    The dissimilarity matrix of the shortest path distances is embedded into $1$-dimension by minimization of raw-stress criterion. 
    In order to test $H_0:\beta=0$ vs $H_1: \beta \neq 0$, the test statistics $F^*$ based on the true regressors
    $t_i$ and $\hat{F}$ based on the $1$-dimensional isomap embeddings $\hat{z}_i$ are comapared, where $n$ is the total number of nodes in the graph. The corresponding powers are empirically estimated by the proportions of times in a
    collection of $100$ Monte Carlo samples the test statistics reject $H_0$, for every $n$ varying from $100$ to $500$ in steps of $25$.
    The plot shows that the difference between the estimated powers of the two tests goes to zero, indicating the tests based on the isomap embeddings are almost as good as the tests based on the true regressors, for sufficiently large number of auxiliary nodes.
    }
    \label{fig 5}
\end{figure}

We assume that the underlying
arclength parameterized 
manifold is $\psi([0,1])$ where $\psi:[0,1] \to \mathbb{R}^4$ is given by
$\psi(t)=(t/2,t/2,t/2,t/2)$. 
We take the number of nodes at which responses are recorded to be $s=20$.
Here, $m$ denotes the number of auxiliary nodes, and $n=m+s$ denotes the total number of nodes. We vary $n$ over the set $\left\lbrace 500,750,1000,.....3000 \right\rbrace$. For each $n$, we repeat the following procedure over $100$ Monte Carlo samples. 
A sample $t_1,....t_s \sim^{iid} U[0,1]$ is generated, and responses $y_i$ are generated from the regression model $y_i=\alpha+\beta t_i +\epsilon_i$, $\epsilon_i \sim^{iid} N(0,\sigma^2_{\epsilon})$ for all $i \in [s]$, where $\alpha=2.0, \beta=5.0, \sigma_{\epsilon}=0.01$.
Additionally, we sample the pre-images of the auxiliary nodes,
$t_{s+1},....t_n \sim^{iid} U[0,1]$. Thus, a random dot product graph with latent positions $\mathbf{x}_i=\psi(t_i)$, $i \in [n]$ is generated and the adjacency spectral estimates $\hat{\mathbf{x}}_i$ of its latent positions are computed.
A localization graph is constructed with the first $n/10$ of the adjacency spectral estimates as nodes under the following rule: two nodes $\hat{\mathbf{x}}_i$ and $\hat{\mathbf{x}}_j$ of the localization graph are to be joined by an edge if and only if
$\left\lVert \hat{\mathbf{x}}_i-\hat{\mathbf{x}}_j \right\rVert<\lambda$, where $\lambda=0.8 \times 0.99^{K-1}$ when $n$ is the $K$-th term in the vector $(500,750,1000, \dots 3000)$.
 Then, the shortest path distance matrix $\mathbf{D}=
 \left(
d_{n,\lambda}(\hat{\mathbf{x}}_i,\hat{\mathbf{x}}_j)
 \right)_{i,j=1}^l
 $ of the first $l=s+1$ estimated latent positions is embedded into one-dimension by raw-stress minimization using $\textrm{mds}$ function of $\textrm{smacof}$ package in R and one-dimensional embeddings $\hat{z}_i$ are obtained. The responses $y_i$ are regressed upon the $1$-dimensional raw-stress embeddings $\hat{z}_i$ for $i \in [s]$, and the predicted response $\Tilde{y}_{s+1}$ for the $(s+1)$-th node  is computed. The predicted  response
$\hat{y}_{s+1}$
for the $(s+1)$-th node based on the true regressors $t_i$ is also obtained. Due to identicality of distributions of the true regressors $t_i$, the distribution of each of the predicted responses is invariant to the label of the node; hence we drop the subscript and use $\hat{y}_{true}$ to denote the predicted response based on the true regressors $t_i$, and use $\hat{y}_{sub}$ to denote the predicted response based on the raw-stress embeddings $\hat{z}_i$ (since raw-stress embeddings are used as substitutes for the true regressors in predicting the response).
 Finally, the sample mean of the squared distances 
 $(\hat{y}_{sub}-\hat{y}_{true})^2$
  over all the Monte Carlo samples is computed and plotted against $n$. The plot is given in \textit{Figure \ref{fig 4}}.
\newline
\newline
\newline
Next, we focus on the issue of hypothesis testing based on the raw-stress embeddings $\hat{z}_i$. In order to test the validity of the model 
\begin{displaymath}
y_i=\alpha+\beta t_i +\epsilon_i
\end{displaymath}
where $\epsilon_i \sim^{iid} N(0,\sigma^2_{\epsilon})$, $i \in [s]$, one would conduct hypothesis testing $H_0: \beta=0$ against $H_1: \beta \neq 0$. If the true regressors $t_i$ were known, we would have used the test statistic $F^*$ of Eq. (\ref{Cor2:stat}), but \textit{Corollary \ref{Cor_test_conv}} tells us that with sufficiently large number of auxiliary latent positions, one can have a  test based on the one-dimensional raw-stress embeddings $\hat{z}_i$, whose power approximates the power of the test based on the true $F$-statistic $F^*$.
We present a plot in \textit{Figure \ref{fig 5}} that speaks in support of \textit{Corollary \ref{Cor_test_conv}}.
We show that the power of the test based on the raw-stress embeddings approaches 
the power of the test based on the true regressors for a chosen value of the pair of the regression parameters. 
 The setting is almost the same as the previous one, except that here the number $n$ of nodes varies from $100$ to $1000$ in steps of $50$.
 For each $n$, the true $F$-statistic (based on the true regressors $t_i$) and the estimated $F$-statistic (based on the raw-stress embeddings $\hat{z}_i$) are computed for $100$ Monte Carlo samples, and
 the power of the two tests are estimated by the proportion of the Monte Carlo samples for which the statistics exceed a particular threshold. 
 Then, for each $n$, the difference between the empirically estimated powers of the two tests (one based on the raw-stress embeddings and the other based on the true regressors) is computed and plotted against $n$. The plot is given in \textit{Figure \ref{fig 5}} which shows that the difference between the empirically estimated powers of the tests based on the true regressors and the raw-stress embeddings approach zero as the number of auxiliary nodes grows.
\newline
\newline
\section{Application} 
\label{Appli}
\begin{figure}[!ht]
    \centering
    \includegraphics{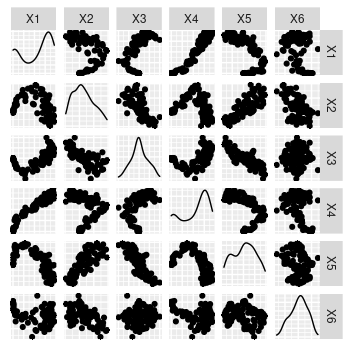}
    \caption{Matrix of scatterplots indicating an underlying low-dimensional structure in the network of $100$ Kenyon Cell neurons in larval Drosophila. A directed graph is taken into account where every node represents a neuron. A $6$-dimensional adjacency spectral estimate is obtained for every node by augmenting the $3$ leading left singular vectors scaled by corresponding singular values, with $3$ leading right singular vectors scaled by corresponding singular values. A scatterplot is then obtained for every pair of these $6$ components. Since each dimension appears to be approximately related to another dimension via a function, presence of an underlying $1$-dimensional structure is assumed.
    }
    \label{fig 6}
\end{figure}

\begin{figure}[!ht]
     \centering
     \includegraphics{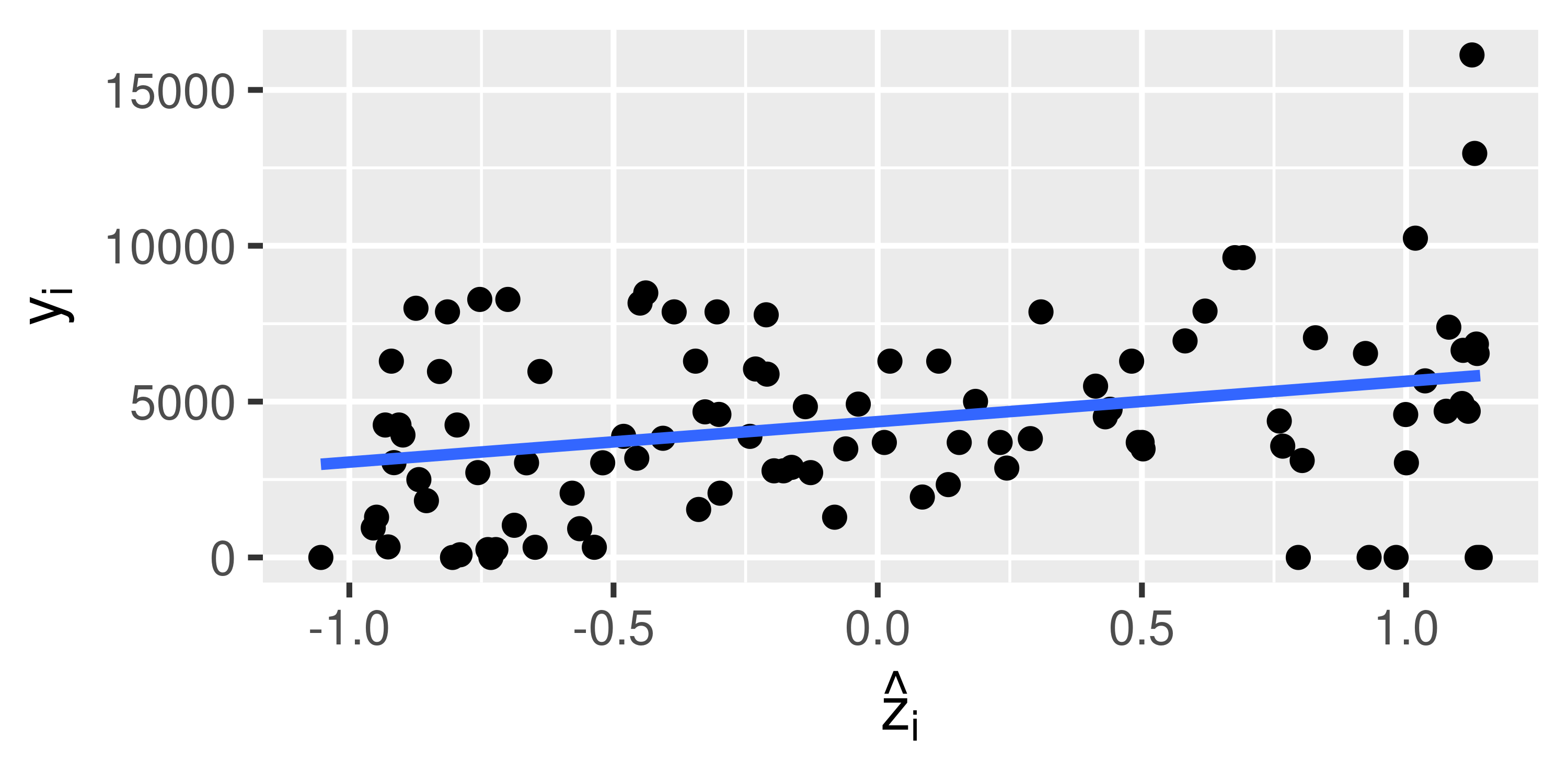}
     \caption{Scatterplot indicating that
     the responses and the $1$-dimensional raw-stress embeddings are linked via a simple linear regression model.
     From $6$-dimensional estimates of the latent positions corresponding to $100$ Kenyon Cell neurons forming a directed network in larval Drosophila, $1$-dimensional embeddings $\hat{z}_i$'s are obtained
     by raw-stress minimization of the shortest path distances. 
     The distance ($y_i$) between bundle entry point of the $i$-th neuron and mushroom body neuropil is treated as the response corresponding to the $i$-th neuron.
     Scatterplot of $(y_i,\hat{z}_i)$,  with fitted regression line $y=4356.1+1296.6x$ indicates a significant effect ($p<0.01$ for $H_0:a=0$ vs $H_1:a \neq 0$ in $y_i=a+b_i \hat{z}_i+ \eta_i$)
     }
     \label{fig 7}
 \end{figure}

In this section, we demonstrate the application of our methodology to real world data. 
Howard Hughes Medical Institute Janelia reconstructed the complete wiring diagram of the higher order parallel fibre system for associative learning in the larval \textit{Drosophila} brain, the mushroom body.
There are $n=100$ datapoints corresponding to $100$ Kenyon cell neurons forming a network in a latent space. 
The distance (in microns) between the bundle entry point of a Kenyon cell neuron and mushroom body neuropil is treated as the response corresponding to that neuron.
We carry out hypothesis testing to test whether a simple linear regression model links the responses on the neurons of the right hemisphere of the larval \textit{Drosophila}  (\cite{Eichler2017TheCC},\cite{https://doi.org/10.48550/arxiv.1705.03297},\cite{athreya2017statistical}) to some dimension-reduced version of the latent positions of the neurons.
\newline
\newline
A directed graph representing a network of the $100$ Kenyon cell neurons is observed. 
Since the graph under consideration is directed, the adjacency spectral embedding is formed by taking into account both the left and right singular vectors of the adjacency matrix. 
The latent position of each node is estimated by a $6$-dimensional vector formed by augmenting the top $3$ left singular vectors scaled by the diagonal matrix of the corresponding singular values, with the top $3$ right singular vectors scaled by the diagonal matrix of the corresponding singular values. 
For each pair of components, we obtain a scatterplot of the bivariate dataset of all $100$ points, thus obtaining a $6 \times 6$ matrix of scatterplots which is shown in
\textit{Figure \ref{fig 6}}.
\newline
\textit{Figure \ref{fig 6}} shows that every component is approximately related to every other component, thus indicating the possibility that the six-dimensional datapoints lie on  a one-dimensional manifold. We construct a localization graph with neighbourhood parameter $\lambda=0.50$ on the $6$-dimensional estimates of the latent positions and embed the dissimilarity matrix 
$
\mathbf{D}=
\left(
d_{100,0.5}(\hat{\mathbf{x}}_i,\hat{\mathbf{x}}_j)
\right)_{i,j=1}^{100}
$ of shortest path distances into one-dimension by minimizing the raw-stress criterion to obtain the $1$-dimensional embeddings $\hat{z}_i$.
 \textit{Figure \ref{fig 7}} presents the plot of responses $y_i$ against the one-dimensional raw-stress embeddings $\hat{z}_i$. 
 \newline
 \newline
The plot in \textit{Figure \ref{fig 7}} 
gives an idea that a simple linear regression model links the responses with the raw-stress embeddings.
We wish to check the validity the model
\begin{displaymath}
y_i=a+b \hat{z}_i + \eta_i
\end{displaymath}
where $\eta_i \sim^{iid} N(0,\sigma^2_{\eta})$, $i \in [n]$. For that purpose, we test $H_0:b=0$ vs $H_1:b \neq 0$ at level of significance $0.01$. The value of the $F$-statistic, with degrees of freedom $1$ and $98$, is found to be $9.815$. This yields 
$p$-value $=P[F_{1,98}>9.815]=0.0023$ which is lower than our level of significance $0.01$. Therefore, we conclude that a simple linear regression model involving $y_i$ as values of the dependent variable and $\hat{z}_i$ as values of the independent variable exist. 
Using \textit{Corollary \ref{Cor_test_conv}}, we conclude that the responses on the neurons are linked to the scalar pre-images of the latent positions via a simple linear regression model. 
 Moreover, if the distances between the bundle entry point and the mushroom body neuropil is not recorded for some Kenyon cell neurons, then the values can be predicted using the one-dimensional raw-stress embeddings $\hat{z}_i$ as proxy for the true regressors.
\newline
\section{Conclusion}
\label{Conc}
\begin{figure}[!ht]
    \centering
    \includegraphics{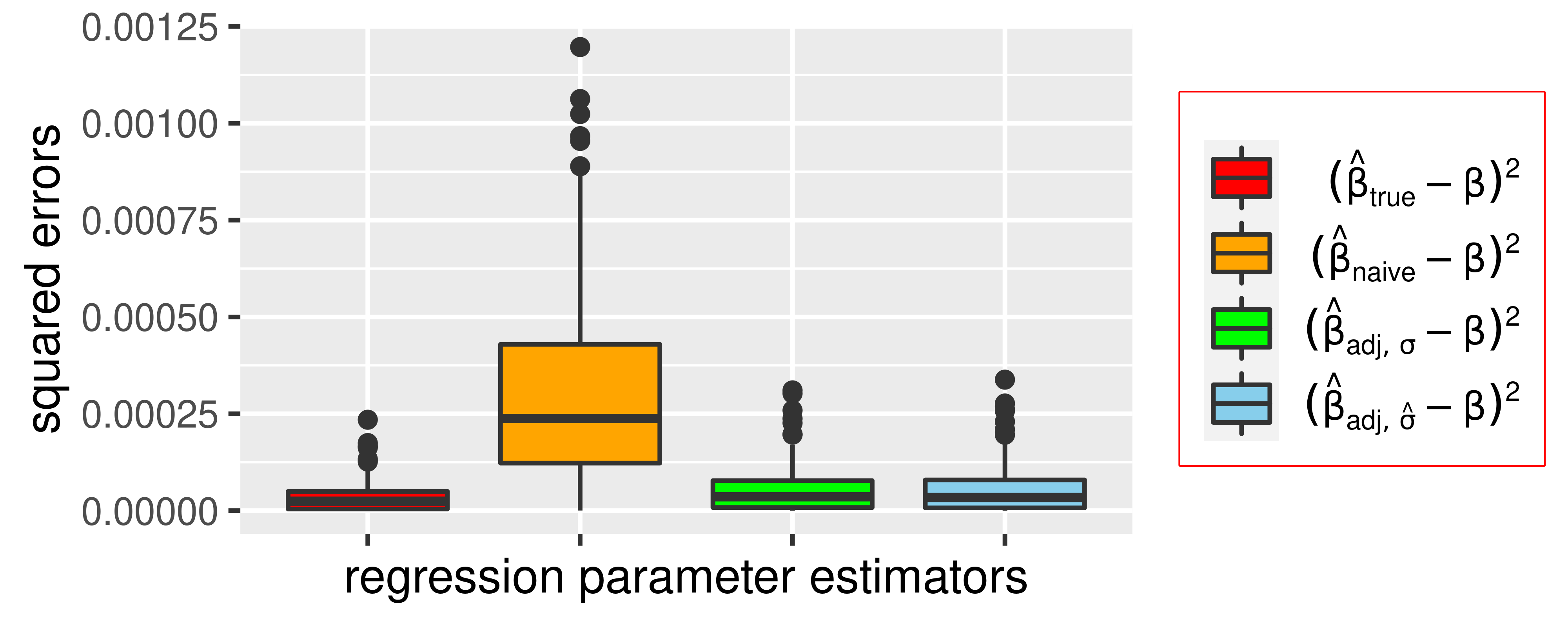}
    \caption{Boxplot of squared errors of the $4$ estimators of the regression slope parameter is given, where the intercept term of the regression model is zero. On each of $100$ Monte Carlo samples, 
    a random graph of $n=800$ nodes is generated for which the latent position of the $i$-th node
    is given by
    $\mathbf{x}_i=(t_i^2, 2t_i(1-t_i),(1-t_i)^2)$ where $t_i \sim^{iid} U[0,1]$.
    Response $y_i$ is generated at the $i$-th node via the regression model $y_i=\beta t_i+\epsilon_i$, $\epsilon_i \sim^{iid} N(0,\sigma^2_{\epsilon})$
    where $\beta=5.0$, $\sigma_{\epsilon}=0.1$.
    The naive estimator was computed by plugging-in the pre-images of the projections of the optimally rotated adjacency spectral estimates of latent positions. In order to compute 
    $\hat{\beta}_{adj,\sigma}$, we plug-in the sum of sample variances obtained from another set of Monte Carlo samples where the graphs are generated from the same model.
    We obtain $\sum_{i=1}^n \hat{\Gamma}_i$, by using delta method on 
    the asymptotic variance (see \ref{eq:Th2st1}) of the optimally rotated adjacency spectral estimates of the latent positions, and thus compute $\hat{\beta}_{adj,\hat{\sigma}}$.
    }
    \label{fig 8}
\end{figure}

In the presented work, theoretical and numerical results are derived on models where latent positions of random dot product graphs lie on a one-dimensional manifold in a high dimensional ambient space. We demonstrated that for a known manifold, the parameters of a simple linear regression model linking the response variable recorded at each node of the graph to the scalar pre-images of the latent positions of the nodes can be estimated consistently even though the true regressors were unknown. However, a key result of our work is to show that even when the manifold is unknown  (the more realistic scenario) one can learn it reasonably well under favourable conditions in order to obtain predicted responses that are close to the predicted responses based on the true regressors. 
\newline
\newline
We use the convergence guarantees for raw-stress embeddings (\cite{trosset2020learning}) to obtain the consistent estimators of the interpoint distances of the regressors when the underlying manifold is unknown. 
We demonstrate that as the number of auxiliary latent positions grow to infinity, at every auxiliary node the predicted response based on the raw-stress embeddings approach the predicted response based on the true regressors. 
\newline
\newline
Observe that while the substitute estimators of the regression parameters  (or the  predicted responses
based on the raw-stress embeddings) can deliver asymptotic performances close to the performance of their counterparts based on the true regressors, in real-life scenarios we can be dealing with small samples, where the substitute estimators (or the predicted responses) are likely to be poor performers. 
When the underlying manifold is known,
we can overcome this issue by taking into account the measurement errors which are the differences between the estimated regressors and the true regressors, thus making adjustments in the estimators of the regression parameters (\cite{fuller1987measurement}).
We conduct a simulation to compare the performances of the estimator based on the true regressors, the substitute (or naive) estimator and the measurement error adjusted estimators, on a known manifold. For a regression model $y_i=\beta t_i + \epsilon_i$,
$\epsilon_i \sim^{iid} N(0,\sigma^2_{\epsilon})$, 
where regressors $t_i$ are estimated by $\hat{t}_i$,
the measurement error adjusted estimator is given by 
$\hat{\beta}_{adj,\sigma}=
\frac
{
\sum_{i=1}^n y_i \hat{t}_i
}
{
\sum_{i=1}^n \hat{t}_i^2 - \sum_{i=1}^n \Gamma_i
}
$
where $\Gamma_i=var(\hat{t}_i-t_i)$. In most realistic scenarios, it is not possible to know the true values of $\Gamma_i$. However, if they admit consistent estimates $\hat{\Gamma}_i$, then we can use the proxy given by
$\hat{\beta}_{adj,\hat{\sigma}}
=
\frac
{
\sum_{i=1}^n y_i \hat{t}_i
}
{
\sum_{i=1}^n \hat{t}_i^2 - \sum_{i=1}^n \hat{\Gamma}_i
}
$. In order to compare the performances of these estimators, we sample a random dot product graph whose nodes lie on a one-dimensional curve in a high dimensional ambient space. We compute the adjacency spectral estimates of the latent positions, project them onto the manifold, and obtain estimates of the regressors which are then used to compute the values of $\hat{\beta}_{true}$, $\hat{\beta}_{naive}$, $\hat{\beta}_{adj,\sigma}$ and $\hat{\beta}_{adj,\hat{\sigma}}$ for $100$ Monte Carlo samples. A boxplot of the values of these estimators computed over $100$ Mont Carlo samples is shown in \textit{Figure \ref{fig 8}}.
\newline
\newline
\textit{Figure \ref{fig 8}} clearly shows that the measurement error adjusted estimators $\hat{\beta}_{adj,\sigma}$ and $\hat{\beta}_{adj,\hat{\sigma}}$ outperform the naive estimator $\hat{\beta}_{naive}$. Moreover, it is also apparent that the performances of  $\hat{\beta}_{adj,\sigma}$ and  
$\hat{\beta}_{adj,\hat{\sigma}}$ don't differ by a significant amount. 
This assures us that the use of measurement error adjusted estimator 
$\hat{\beta}_{adj,\hat{\sigma}}$ is pragmatic and effective,
since the computation of $\hat{\beta}_{adj,\sigma}$ is not possible in many realistic scenarios owing to the lack of knowledge of the true values of $\Gamma_i$. 
\newline
\newline
Unfortunately, in the case of unknown manifolds, one cannot readily apply this same methodology, as only interpoint distances, and not embeddings are preserved, as discussed in \textit{Section \ref{Theoretical_Results}}.
 We believe it can be an interesting problem to approach in future. Another area of future investigation can be to see whether the results for one-dimensional manifold can be generalized to $k$-dimensional manifolds where $k>1$. 
\newpage


\begin{backmatter}
\section*{Availability of data and materials}
The dataset analyzed in this study are included in the published article \cite{Eichler2017TheCC}.

\section*{Competing interests}
  The authors declare that they have no competing interests.

\section*{Author's contributions}
    AA developed the theory, conducted experiments and wrote the manuscript.  JA developed the theory and edited the manuscript. MWT developed the theory and edited the manuscript.
    YP conducted experiments and edited the manuscript. CEP formulated the problem, developed the theory and edited the manuscript.

\section*{Acknowledgements}
  This work is partially supported by the Johns Hopkins Mathematical Institute for Data Science (MINDS) Fellowship. 

\bibliographystyle{bmc-mathphys} 
\nocite{alyakin2020correcting}
\nocite{athreya2016limit}
\nocite{athreya2017statistical}
\nocite{athreya2021estimation}
\nocite{bernstein2000graph}
\nocite{Cape2019TheTN}
\nocite{fuller1987measurement}
\nocite{https://doi.org/10.48550/arxiv.1409.2344}
\nocite{https://doi.org/10.48550/arxiv.1705.03297}
\nocite{https://doi.org/10.48550/arxiv.1903.08656}
\nocite{rubin2020manifold}
\nocite{sekhon2021result}
\nocite{Eichler2017TheCC}
\nocite{goldenberg2010survey}
\nocite{Erdos1984OnTE}
\nocite{Hoff2002LatentSA}
\nocite{Young2007RandomDP}
\nocite{1326716}
\nocite{6789755}
\nocite{borg2005modern}
\nocite{trosset2021rehabilitating}
\bibliography{bmc_article}
\end{backmatter}
\newpage
\section{Appendix}
\label{Appendix}
\textit{
\textbf{Theorem 4:}  Suppose $\psi:[0,L] \to \mathbb{R}^d$ is bijective, and its inverse $\gamma$ satisfies $\left\lVert \nabla\gamma(\mathbf{w}) 
\right\rVert<K$ 
for all $\mathbf{w} \in \psi([0,L])$,
for some $K>0$. Let $\mathbf{x}_i=\psi(t_i)$ be the latent position of the $i$-th node of a random dot product graph with $n$ nodes, and assume $y_i=\alpha+\beta t_i +\epsilon_i$, $\epsilon_i \sim^{iid} N(0,\sigma^2_{\epsilon})$ for all $i \in [n]$. Assume $\mathbf{x}_i \sim^{iid} F$ for all $i$ where $F$ is an inner product distribution on $\mathbb{R}^d$.
Suppose $\Tilde{\mathbf{x}}_i$ is the optimally rotated adjacency spectral estimate of $\mathbf{x}_i$ for all $i$, and 
\newline
$\hat{t}_i=\arg \min_t \left\lVert \Tilde{\mathbf{x}}_i-\psi(t) \right\rVert$.
Then, $\hat{\alpha}_{sub} \to^P \alpha$, $\hat{\beta}_{sub} \to^P \beta$ as $n \to \infty$.
\newline
\newline
\textbf{Proof:} 
Set $\mathbf{u}_i=\Tilde{\mathbf{x}}_i-\mathbf{x}_i$ for all $i \in [n]$ and
note that by \textit{Theorem 1}, 
$\max_i \left\lVert \mathbf{u}_i \right\rVert \to^P 0$ as $n \to \infty$.
Let $\mathbf{u}_i=\Tilde{\mathbf{x}}_i-\mathbf{x}_i$ and let $\mathbf{h}_i$ be the vector of minimum length for which $\Tilde{\mathbf{x}}_i+\mathbf{h}_i \in \psi([0,L])$. Note that 
$\left\lVert \mathbf{h}_i \right\rVert \leq \left\lVert 
\mathbf{u}_i \right\rVert$ for all $i$.
\newline
Setting $\mathbf{q}_i=\mathbf{h}_i+\mathbf{u}_i$ and using Taylor's theorem, we observe that for all $i$,
\begin{displaymath}
\begin{aligned} 
\hat{t}_i=\gamma(\mathbf{x}_i+\mathbf{q}_i)=t_i+ \mathbf{q}_i^T \nabla \gamma(\mathbf{x}_i)+
    o(\left\lVert \mathbf{q}_i \right\rVert).
\end{aligned}
\end{displaymath}
Hence, by Cauchy-Schwarz Inequality, 
\begin{displaymath}
  |\hat{t}_i-t_i| \leq \left\lVert \mathbf{q}_i \right\rVert \left\lVert \nabla\gamma(\mathbf{\mathbf{x}}_i) \right\rVert
+o(\left\lVert \mathbf{q}_i \right\rVert)
\leq K\left\lVert \mathbf{q}_i \right\rVert +
o(\left\lVert \mathbf{q}_i \right\rVert). 
\end{displaymath}
\newline
Note that $\left\lVert \mathbf{q}_i \right\rVert \leq 2 \left\lVert \mathbf{u}_i \right\rVert$ by Triangle Inequality, and therefore $\max_i \left\lVert \mathbf{q}_i \right\rVert \to^P 0$ as $n \to \infty$, which implies $\max_i |\hat{t}_i-t_i| \to^P 0$ as $n \to \infty$.
\newline
Recall that the regression parameter estimators based on true regressor values are
\begin{displaymath}
    \hat{\beta}_{true}=
    \frac
    { \frac{1}{n}
     \sum_{i=1}^n (y_i-\Bar{y})(t_i-\Bar{t})
    }
    { \frac{1}{n}
    \sum_{i=1}^n (t_i-\Bar{t})^2
    }, \hspace{0.5cm}
    \hat{\alpha}_{true}=
    \Bar{y}-\hat{\beta}_{true} \Bar{t},
\end{displaymath}
and the substitute or plug-in estimators are
\begin{displaymath}
     \hat{\beta}_{sub}=
    \frac
    { \frac{1}{n}
     \sum_{i=1}^n (y_i-\Bar{y})(\hat{t}_i-\Bar{\hat{t}})
    }
    { \frac{1}{n}
    \sum_{i=1}^n (\hat{t}_i-\Bar{\hat{t}})^2
    }, \hspace{0.5cm}
    \hat{\alpha}_{sub}=
    \Bar{y}-\hat{\beta}_{sub} \Bar{\hat{t}}.
\end{displaymath}
Note that by Triangle Inequality, as $n \to \infty$, $\max_i |\hat{t}_i-t_i| \to^P 0$ implies $|\Bar{t}-\Bar{\hat{t}}| \to^P 0$, $|\frac{1}{n}
    \sum_{i=1}^n (t_i-\Bar{t})^2
    -
    \frac{1}{n}
    \sum_{i=1}^n (\hat{t}_i-\Bar{\hat{t}})^2| \to^P 0
    $ and
$
|\frac{1}{n}
     \sum_{i=1}^n y_i(t_i-\Bar{t})-
\frac{1}{n}
     \sum_{i=1}^n y_i(\hat{t}_i-\Bar{\hat{t}})|
\to^P 0
$.
Thus, as $n \to \infty$, $|\hat{\beta}_{sub}-\hat{\beta}_{true}| \to^P 0$ and $|\hat{\alpha}_{sub}-\hat{\alpha}_{true}| \to^P 0$. Recalling $\hat{\alpha}_{true}$ and $\hat{\beta}_{true}$ are consistent for $\alpha$ and $\beta$ respectively, we conclude $\hat{\alpha}_{sub}$ and $\hat{\beta}_{sub}$ too are consistent for $\alpha$ and $\beta$.
}
\newline
\newline
\newline
\textit{
\textbf{Corollary 1:}
Conditioning upon the true regressors $t_i$ in the setting of \textit{Theorem \ref{Th_kn_const}},
the following two conditions hold
\newline
(A)
$
\mathbb{E}(\hat{\alpha}_{sub}) \to \alpha,
\hspace{0.5cm} 
\mathbb{E}(\hat{\beta}_{sub}) \to \beta 
\text{ as $n \to \infty$}.
$,
\newline
(B)
For any two linear unbiased estimators $\Tilde{\alpha}$ and $\Tilde{\beta}$ and an arbitrary $\delta>0$, 
\newline
$
\mathrm{var}(\hat{\alpha}_{sub}) \leq \mathrm{var}(\Tilde{\alpha})+\delta, \hspace{0.5cm}
\mathrm{var}(\hat{\beta}_{sub}) \leq \mathrm{var}(\Tilde{\beta})+\delta
$ for sufficiently large $n$.
\newline
\newline
\textbf{Proof:} From \textit{Theorem \ref{Th_kn_const}}
it directly follows that as $n \to \infty$, 
$E(\hat{\alpha}_{sub}) \to \alpha$,
$E(\hat{\beta}_{sub}) \to \beta$.
Moreover, note that $(\hat{\alpha}_{sub}-\hat{\alpha}_{true}) \to^P 0$ and $(\hat{\beta}_{sub}-\hat{\beta}_{true}) \to^P 0$ as $n \to \infty$. Thus, for any $\delta>0$, $var(\hat{\alpha}_{sub}) \leq var(\hat{\alpha}_{true})+\delta$,
$var(\hat{\beta}_{sub}) \leq var(\hat{\beta}_{true}) + \delta$ for sufficiently large $n$. Recalling that $\hat{\alpha}_{true}$ and $\hat{\beta}_{true}$ are best linear unbiased estimators of $\alpha$ and $\beta$ respectively, $(B)$ follows.
}
\newline
\newline
\textit{
\textbf{Theorem 6:}  Consider a  random dot product graph for which each node lies on an arclength parameterized one-dimensional manifold $\psi([0,L])$ where $\psi$ is unknown.
Let $\mathbf{x}_i=\psi(t_i)$ be the latent position of the $i$-th node for all $i$. Assume  $y_i=\alpha+\beta t_i + \epsilon_i$, $\epsilon_i \sim^{iid} N(0,\sigma^2_{\epsilon})$ for $i \in [s]$,
where $s$ is a fixed integer.
The predicted response at the $r$-th node based on the true regressors is 
$\hat{y}_r=\hat{\alpha}_{true}+\hat{\beta}_{true} t_r$. 
There exist sequences $n_K \to \infty$ of number of nodes and $\lambda_K \to 0$ of neighbourhood parameters such that for every $r>s$, 
$
|\hat{y}_r-\Tilde{y}_r^{(K)}| \to^P 0
$
as $K \to \infty$,
where
$\Tilde{y}_r^{(K)}=\mathrm{PRED}(\mathbf{A}^{(K)},d,\lambda_K,l,
\left\lbrace y_i \right\rbrace_{i=1}^s, r
)$ (see \textit{Algorithm \ref{Algo2unknown}}), $\mathbf{A}^{(K)}$ being the adjacency matrix when the number of nodes is $n_K$ and $l$ being a fixed natural number that satisfies $l>r>s$.
\newline
\newline
\textbf{Proof:} 
Fix $l \in \mathbb{N}$ such that $s<r \leq l$. 
For each $K \in \mathbb{N}$, choose number of nodes $n_K$ to be observed and appropriate $\lambda_K$ such that eqn
\ref{eq:Th4st1} holds, and recall from \textit{eqn} \ref{eq:Th4st2} that
 $(\hat{z}_1^{(K)},....\hat{z}_l^{(K)})$ is the minimizer of the raw stress criterion:
\begin{equation*}
    \sigma_l(z_1,...z_l)=
    \frac{1}{2}
    \sum_{i=1}^l \sum_{j=1}^l (|z_i-z_j|-d_{n_K,\lambda_K}(\hat{\mathbf{x}}_i,\hat{\mathbf{x}}_j))^2
\end{equation*}
From \textit{Theorem 5}, we know that for all $i,j \in [l]$, as $K \to \infty$,
\begin{equation}
(|\hat{z}_i^{(K)}-\hat{z}_j^{(K)}|-|t_i-t_j|) \to^P 0.
\end{equation}
Define
\begin{equation}
\Tilde{\beta}^{(K)}=
\frac
{
\sum_{i=1}^s (y_i-\Bar{y})(\hat{z}_i^{(K)}-\Bar{\hat{z}}^{(K)})
}
{
\sum_{i=1}^s (\hat{z}_i^{(K)}-\Bar{\hat{z}}^{(K)})^2
}, \hspace{1cm}
\Tilde{\alpha}^{(K)}=
\Bar{y}-\Tilde{\beta}^{(K)} \Bar{\hat{z}}^{(K)}
\label{eqn:20}
\end{equation}
where $\Bar{\hat{z}}^{(K)}=\frac{1}{s} \sum_{i=1}^s \hat{z}_i^{(K)}$.
Then we can define the predictor of $y_r$ based on $\hat{z}_i^{(K)}$'s to be
\begin{displaymath}
\Tilde{y}_r^{(K)}=\Tilde{\alpha}^{(K)}+\Tilde{\beta}^{(K)} \hat{z}_r^{(K)}=\Bar{y}+\Tilde{\beta}^{(K)} (\hat{z}_r^{(K)}-\Bar{\hat{z}}^{(K)}).
\end{displaymath}
If original $t_i$'s were known, the predictor of $y_r$ would be 
\begin{displaymath}
\hat{y}_r=\hat{\alpha}+\hat{\beta} t_r=\Bar{y}+\hat{\beta}(t_r-\Bar{t}).
\end{displaymath}
Now, recall that for all $i,j \in [l]$,
$(|\hat{z}_i^{(K)}-\hat{z}_j^{(K)}|-|t_i-t_j|) \to^P 0$
as $K \to \infty$. 
Thus, for any $\tau>0$ and $\nu>0$, there exists $K_0 \in \mathbb{N}$ such that
for all $K \geq K_0$, with probability at least $(1-\nu)$, for all
$i \in [l]$,
\begin{equation}
\begin{aligned} 
&|a^{(K)}t_i+b^{(K)}- \hat{z}_i^{(K)}|
\leq \tau
&\implies
|a^{(K)}(t_i-\Bar{t})-
(\hat{z}_i^{(K)}-\Bar{\hat{z}}^{(K)})|
\leq  2\tau
\end{aligned}
\end{equation}
where $a^{(K)} \in \left\lbrace -1,+1 \right\rbrace$,
$b^{(K)} \in \mathbb{R}$.
Note that $(a^{(K)})^2=1$ for all $K$. Thus, taking sufficiently large $K$, we can bring $\sum_{i=1}^s y_i(t_i-\Bar{t})(t_r-\Bar{t})$ and
$\sum_{i=1}^s y_i(\hat{z}_i^{(K)}-\Bar{\hat{z}}^{(K)})(\hat{z}^{(K)}_r-
\Bar{\hat{z}}^{(K)})$ arbitrarily close with arbitrarily high probability.
We can also bring $\sum_{i=1}^s (\hat{z}_i^{(K)}-\Bar{\hat{z}}^{(K)})^2$
and $\sum_{i=1}^s (t_i-\Bar{t})^2$ arbitrarily close with arbitrarily high probability, by choosing sufficiently large $K$. 
Recall that
\begin{equation}
\begin{aligned}
\Tilde{y}_r^{(K)}=
\Bar{y}+
\frac
{
\sum_{i=1}^s y_i(\hat{z}_i^{(K)}-\Bar{\hat{z}}^{(K)})(\hat{z}^{(K)}_r-
\Bar{\hat{z}}^{(K)})
}
{
\sum_{i=1}^s (\hat{z}_i^{(K)}-\Bar{\hat{z}}^{(K)})^2
},  \\
\hat{y}_r=\Bar{y}+
\frac
{
\sum_{i=1}^s y_i(t_i-\Bar{t})(t_r-\Bar{t})
}
{
\sum_{i=1}^s (t_i-\Bar{t})^2
}.
\end{aligned}
\end{equation}
Thus, we can bring $\hat{y}_r$ and $\Tilde{y}_r^{(K)}$ arbitrarily close with arbitrarily high probability, by choosing sufficiently large $K$,
which means $|\Tilde{y}_r^{(K)}-\hat{y}_r| \to^P 0$ as $K \to \infty$.
}
\newline
\newline
\textit{
\textbf{Corollary 2:} 
In the setting of \textit{Theorem \ref{Th_reg_pred_const}},
suppose $\left\lbrace (\Tilde{y}_1^{(K)},\Tilde{y}_2^{(K)},....
\Tilde{y}_s^{(K)})
\right\rbrace_{K=1}^{\infty}$ is the sequence of vector of predicted responses at the first $s$ nodes of the random dot product graph, based on the raw-stress embeddings $\hat{z}_1,...,\hat{z}_s$.
Define
\begin{equation}
F^*=
(s-2)
\frac
{
\sum_{i=1}^s (\hat{y}_i-\Bar{y})^2
}
{
\sum_{i=1}^s (y_i-\hat{y}_i)^2
},
\hspace{1cm}
\hat{F}^{(K)}=
(s-2)
\frac
{
\sum_{i=1}^s (\Tilde{y}^{(K)}_i-\Bar{y})^2
}
{
\sum_{i=1}^s (y_i-\Tilde{y}^{(K)}_i)^2
}.
\end{equation}
Consider testing the null hypothesis 
$H_0:\beta=0$ against $H_1:\beta \neq 0$ in the absence of the true regressors $t_i$, and the decision rule is: reject $H_0$ in favour of $H_1$ at level of significance $\Tilde{\alpha}$ if
$\hat{F}^{(K)}>c_{\Tilde{\alpha}}$,
where $c_{\Tilde{\alpha}}$ is the $(1-\Tilde{\alpha})$-th quantile of $F_{1,s-2}$ distribution.
If the power of this test is denoted by $\hat{\pi}^{(K)}$, then $\lim_{K \to \infty} \hat{\pi}^{(K)}=\pi^*$, where $\pi^*$ is the power of the test for which the decision rule is to reject $H_0$ in favour of $H_1$ at level of significance $\Tilde{\alpha}$ if $F^*>c_{\Tilde{\alpha}}$.
\newline
\newline
\textbf{Proof:} 
From \textit{Theorem 5}, we have
$\max_{i \in [s]} |\Tilde{y}_i^{(K)}-\hat{y}_i| \to^P 0$, which implies
$(\hat{F}^{(K)}-F^*) \to^P 0$ as $K \to \infty$.
Thus, for any $(\alpha,\beta) \in \mathbb{R}^2$, as $K \to \infty$,
$P_{\alpha,\beta}[\hat{F}^{(K)}>c_{\Tilde{\alpha}}] \to 
P_{\alpha,\beta}[F^*>c_{\Tilde{\alpha}}]$.
}



\end{document}